%% file: TIP-final.tex
\documentclass[lettersize,journal,review]{IEEEtran}

\usepackage{graphicx}
\usepackage{amssymb,amsmath,epsfig}
\usepackage{algorithm}
\usepackage{algorithmic}
\usepackage{booktabs}
\usepackage{mathrsfs}
\usepackage{color}
\usepackage{textcomp}
\usepackage{bbding}
\usepackage{pifont}
\usepackage{wasysym}
\usepackage{amssymb}
\usepackage{subcaption}
\usepackage{cite}
\usepackage{amsmath,amssymb,amsfonts}
\usepackage{xcolor,colortbl}
\usepackage{overpic}
\usepackage{arydshln}
\usepackage{microtype}
\usepackage{multirow}
\usepackage{makecell}
\usepackage{hhline}
\usepackage[american]{babel}
\usepackage{cuted}
\usepackage{ragged2e}
\usepackage[export]{adjustbox}
\usepackage{array}

\definecolor{lightgray}{gray}{.92}
\definecolor{tinygray}{gray}{.96}

\newcommand{\ie}{\textit{i}.\textit{e}.}
\newcommand{\eg}{\textit{e}.\textit{g}.}

\usepackage{multirow}
\usepackage{epsfig}
\usepackage{adjustbox}
\usepackage[pagebackref=true,
            breaklinks=true,colorlinks,
            citecolor=cyan,linkcolor=red,
            bookmarks=false]{hyperref}

\ifCLASSINFOpdf
\else
\fi
\usepackage[switch]{lineno}

\begin{document}

\title{Transformer-Progressive Mamba Network for Lightweight Image Super-Resolution}

\author{Sichen Guo,
        Wenjie Li,
        Yuanyang Liu,
        Guangwei Gao,~\IEEEmembership{Senior Member,~IEEE,}\\
        Jian Yang,~\IEEEmembership{Member,~IEEE,}
        and Chia-Wen Lin,~\IEEEmembership{Fellow,~IEEE}
  \IEEEcompsocitemizethanks{\IEEEcompsocthanksitem This work was supported in part by the Foundation of the State Key Laboratory of Integrated Services Networks of Xidian University under Grant No. ISN27-4, and also supported by the National Natural Science Foundation of China under Grant Nos. U24A20330 and 62361166670.~\textit{(Sichen Guo and Wenjie Li contributed equally to this work.) (Corresponding author: Guangwei Gao.)}}
\IEEEcompsocitemizethanks{\IEEEcompsocthanksitem Sichen Guo and Yuanyang Liu are with the Bell Honors School, Nanjing University of Posts and Telecommunications, Nanjing 210023, China, and also with the State Key Laboratory of Integrated Services Networks, Xidian University, Xi’an 710071, China (e-mail: \{q22010218, q22010108\}@njupt.edu.cn).}
\IEEEcompsocitemizethanks{\IEEEcompsocthanksitem Wenjie Li is with the Pattern Recognition and Intelligent System Laboratory, School of Artificial Intelligence, Beijing University of Posts and Telecommunications, Beijing 100080, China (e-mail: lewj2408@gmail.com).}
\IEEEcompsocitemizethanks{\IEEEcompsocthanksitem Guangwei Gao and Jian Yang are with the PCA Lab, Key Lab of Intelligent Perception and Systems for High-Dimensional Information of Ministry of Education, School of Computer Science and Engineering, Nanjing University of Science and Technology, Nanjing 210094, China, and Guangwei Gao is also with the State Key Laboratory of Integrated Services Networks, Xidian University, Xi’an 710071, China (e-mail: \{gwgao, csjyang\}@njust.edu.cn).}
\IEEEcompsocitemizethanks{\IEEEcompsocthanksitem Chia-Wen Lin is with the Department of Electrical Engineering and the Institute of Communications Engineering, National Tsing Hua University, Hsinchu 300044, Taiwan (e-mail: cwlin@ee.nthu.edu.tw).}
}

\markboth{IEEE Transactions on Image Processing}%
{Shell \MakeLowercase{\textit{et al.}}: A Sample Article Using IEEEtran.cls for IEEE Journals}

\maketitle

\begin{abstract}
Recently, Mamba-based super-resolution (SR) methods have demonstrated the ability to capture global receptive fields with linear complexity, addressing the quadratic computational cost of Transformer-based SR approaches. However, existing Mamba-based methods lack fine-grained transitions across different modeling scales, which limits the efficiency of feature representation. In this paper, we propose T-PMambaSR, a lightweight SR framework that integrates window-based self-attention with Progressive Mamba. By enabling interactions among receptive fields of different scales, our method establishes a fine-grained modeling paradigm that progressively enhances feature representation without introducing additional computational cost. Furthermore, we introduce an Adaptive High-Frequency Refinement Module (AHFRM) to recover high-frequency details lost during Transformer and Mamba processing. Extensive experiments demonstrate that T-PMambaSR progressively enhances the model's receptive field and expressiveness, achieving competitive performance with recent Transformer- or Mamba-based methods while incurring lower computational cost. Codes will be available at \url{https://github.com/IVIPLab/T-PMambaSR}.
\end{abstract}

\begin{IEEEkeywords}
Lightweight Super-Resolution, Mamba-based, Receptive Field
\end{IEEEkeywords}

\IEEEpeerreviewmaketitle

\section{Introduction}
\label{introduction}

\IEEEPARstart{L}{IGHTWEIGHT} image super-resolution (SR)~\cite{li2024efficient} aims to reconstruct a high-resolution (HR) image from a low-resolution (LR) input with limited computational cost, and has broad applications in image processing~\cite{peng2024towards}, video enhancement~\cite{feng2025pmq}, and security surveillance~\cite{li2025self}. Early CNN-based methods~\cite{gao2022feature,li2025survey} are constrained by their inherently local receptive fields. Even with increased depth, CNNs still struggle to capture long-range dependencies, which are critical for accurately restoring global structures and fine textures. With the development of Vision Transformers~\cite{dosovitskiy2021an}, Transformer-based SR methods~\cite{liang2021swinir,zhang2022efficient,zhou2023srformer,zhang2024hit,li2025dual} have addressed this limitation by leveraging self-attention to model long-range dependencies. However, the quadratic computational complexity of the above Transformer-based SR methods makes them prohibitively expensive for lightweight SR scenarios.

\begin{figure}[tbp]   
  \centering
  \includegraphics[width=\linewidth]{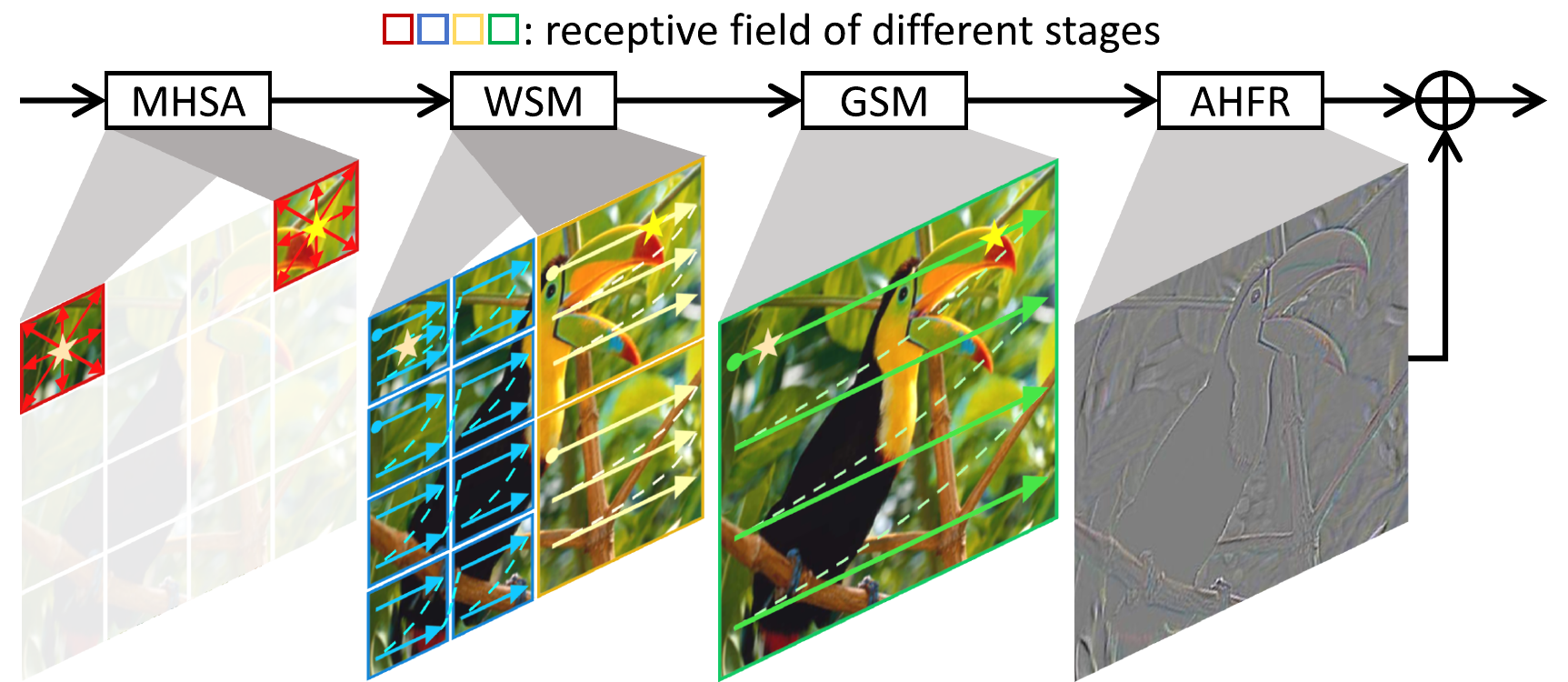} 
  \includegraphics[width=\linewidth]{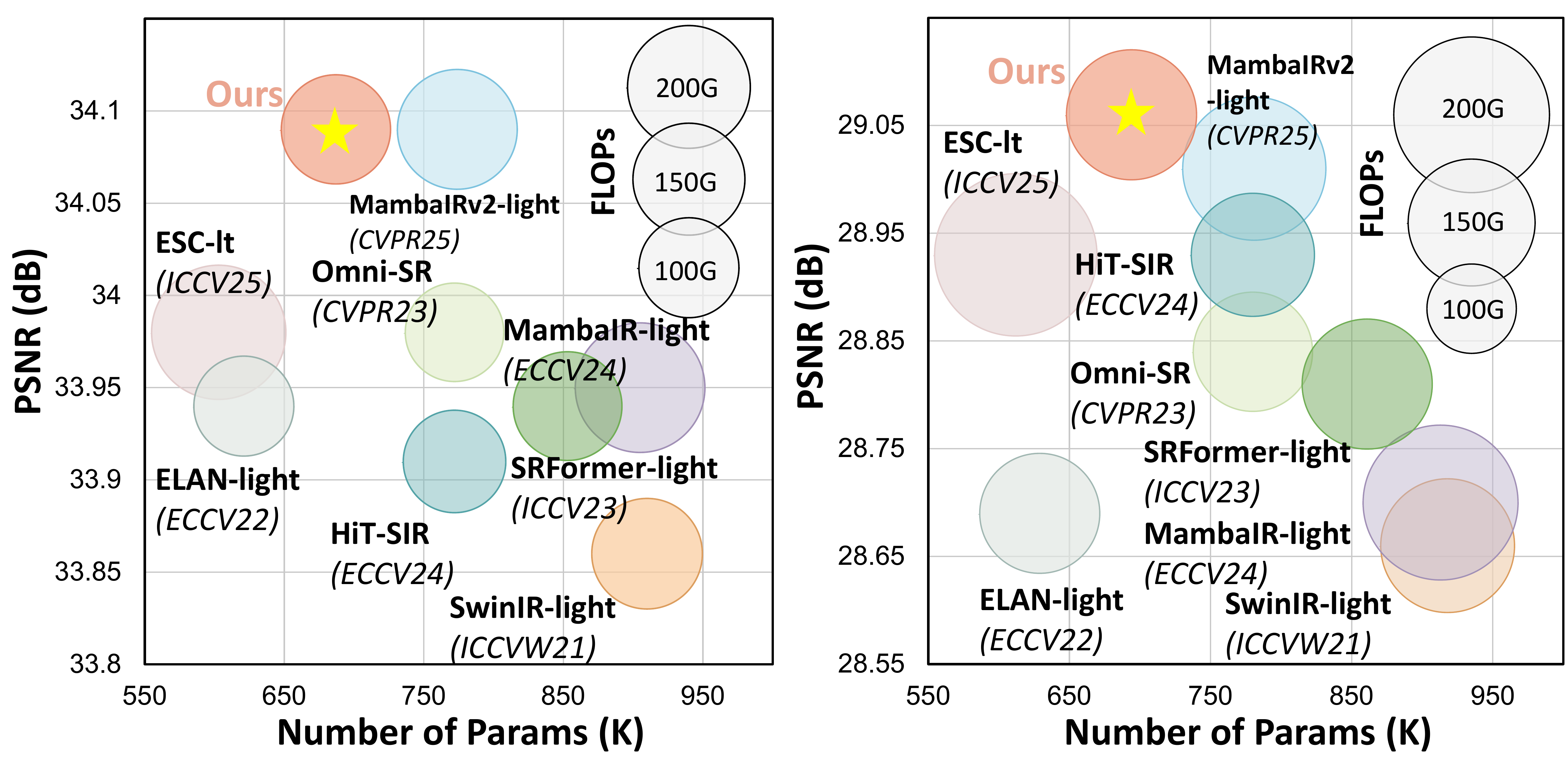}
  \put(-250,-9){\color{black}{\fontsize{7pt}{1pt}\selectfont (1) Trade-off on $\times 2$ Set14~\cite{zeyde2010single} test set.}}
  \put(-125,-9){\color{black}{\fontsize{7pt}{1pt}\selectfont (2) Trade-off on $\times 3$ Urban100~\cite{huang2015single} test set.}}

  \vspace{2mm}
  \caption{(\textbf{Top}): Our design rationale is based on progressively exploiting internal interactions within Window Multi-Head Self-Attention (W-MHSA), combined with a Window Scan Mamba (WSM) and a Global Scan Mamba (GSM) operation. This hierarchical structure facilitates the gradual expansion of receptive fields, ensuring comprehensive information exchange both within and across windows. (\textbf{Bottom}): Leveraging our design, our method strikes an optimal balance across Params, FLOPs, and PSNR, surpassing existing Transformer- and Mamba-based methods.}
  \label{fig:motivation}
\end{figure}

Mamba-based lightweight SR methods~\cite{guo2024mambair,guo2025mambairv2} have shown that state space models (SSMs)~\cite{gu2024mamba} can effectively capture global context with a complexity linear to the spatial resolution, surpassing Transformer-based baselines~\cite{liang2021swinir,zhou2023srformer,li2025dual}. However, existing Mamba-based approaches~\cite{guo2024mambair,guo2025mambairv2} often integrated components with different receptive fields in a coarse-grained manner. For instance, MambaIRv2~\cite{guo2025mambairv2} employed window attention and SSM modules as separate stages, resulting in hierarchical discontinuity: the model transitions abruptly from the fine-grained local details captured by window attention to the broad global context modeled by Mamba. This lacks a progressive mechanism for cross-scale feature fusion, limiting the ability to synergize local precision with global scope and ultimately leading to additional computational overhead.

To address this challenge, we introduce Transformer-Progressive Mamba (T-PMambaSR), a lightweight SR architecture that effectively combines multi-level feature modeling. As illustrated at the top of Fig.~\ref{fig:motivation}, the design integrates standard Window Multi-Head Self-Attention (W-MHSA) with two specialized state-space modules. First, the Window Scan Mamba Layer (WSML) enables both intra-window modeling and long-range inter-window interactions through multi-scale window scanning, progressively expanding the modeling windows and enriching the feature representation with multi-scale information. Subsequently, the Global Scan Mamba Layer (GSML) captures global image features from multiple directions via a computationally efficient multi-head mechanism, maintaining high performance without degradation. The MHSA → WSML → GSML framework forms a hierarchical network that incrementally expands the receptive field, effectively overcoming the locality constraints of window attention and ensuring smooth information flow across both local, regional, and global scales.

Moreover, to mitigate the loss of high-frequency details in SR, we introduce an Adaptive High-Frequency Refinement Module (AHFRM). Prior studies~\cite{li2023feature,xu2025fadpnet} have demonstrated that although Transformer- and Mamba-based architectures excel at capturing low-frequency information, they are inherently limited in modeling high-frequency representations. As features propagate through these low-pass-filter-like modules, high-frequency information is gradually attenuated. To address this issue, AHFRM employs a high-frequency filter to extract high-frequency features from both the input and output of the Transformer–Mamba block. The preserved pre-processed features are then used to guide the recovery of degraded post-processed content. By embedding high-frequency recovery into the network, our T-PMambaSR enhances the sharpness of SR images without compromising efficiency. Based on the above model architectures, as illustrated at the bottom of Fig.~\ref{fig:motivation}, our T-PMambaSR achieves highly efficient SR, offering favorable trade-offs among model size and reconstruction quality. 

 In summary, the contributions of this paper are as follows:

\begin{itemize}
  \item Our T-PMambaSR gradually expands the receptive field, enabling hierarchical feature modeling to transition from local to global in a fine-grained manner, thereby improving the efficiency in capturing features at different levels.
  \item Our AHFRM counteracts the low-pass filtering tendencies inherent in both Transformer and Mamba modules, specifically recovering high-frequency contexts and edges that are lost to low-pass filtering.
  \item Extensive experiments show that our T-PMambaSR achieves competitive performance compared with most existing lightweight Transformer- and Mamba-based SR methods, both qualitatively and quantitatively, with less computational cost.
\end{itemize}

\vspace{-3mm}
\section{Related Work}~\label{Related}
\vspace{-8mm}

\subsection{Lightweight SR Methods}
The development of lightweight SR~\cite{zhao2024efficient} has evolved through several architectural paradigms. Early efforts were dominated by Convolutional Neural Networks (CNNs), where foundational works established key principles for efficiency. For instance, FSRCNN~\cite{dong2016accelerating} pioneered processing features in the LR space to reduce computational cost, while FDIWN~\cite{gao2022feature} further refined network design with sophisticated block-level techniques such as information distillation. These models excelled at local feature representation but were inherently limited by their local receptive fields. To capture long-range dependencies, Vision Transformers were adapted for SR. For example, SwinIR~\cite{liang2021swinir} introduced a highly effective window-based approach to manage the quadratic complexity of self-attention. To overcome the limited receptive field of fixed windows in windows-based SR methods~\cite{gao2022lightweight,li2023cross}, SRFormer~\cite{zhou2023srformer} introduced permuted self-attention to enable larger window sizes without prohibitive cost, while HiT-SIR~\cite{zhang2024hit} employed a hierarchical design with progressively expanding windows to capture multi-scale context. However, to mitigate the quadratic computational complexity associated with increasing window sizes, these approaches often compromise spatial structural information: HiT-SIR relies on lossy spatial downsampling, while SRFormer alters spatial structures via its permuted mechanism. Furthermore, these Transformer-based methods generally lack effective long-range inter-window interactions. 

More recently, works based on State Space Models (SSMs)~\cite{guo2024mambair} have emerged as a powerful alternative for SR, offering global receptive fields with linear complexity. Recent works like MambaIRv2~\cite{guo2025mambairv2} demonstrated the powerful performance of the hybrid Transformer-Mamba architecture. Similarly, SRMamba-T~\cite{liu2025srmamba} explored Mamba-Transformer hybridization, while HiMamba~\cite{qiao2024hi} introduced hierarchical state-space modeling. However, these architectures typically stack modules in a coarse-grained manner and lack mechanisms to smoothly bridge discrete features at different scales, which can lead to abrupt receptive field transitions. 

In contrast, our T-PMambaSR addresses the limitations of both paradigms by establishing a strictly \textbf{fine-grained transition} that progresses smoothly from local to global scales without any spatial downsampling. By altering window-based selective scan paths via our WSML and GSML modules rather than enlarging attention windows, we achieve efficient inter-window communication and progressive receptive field expansion without incurring additional computational overhead.

\begin{figure*}[t]   
  \centering
  \includegraphics[width=\linewidth]{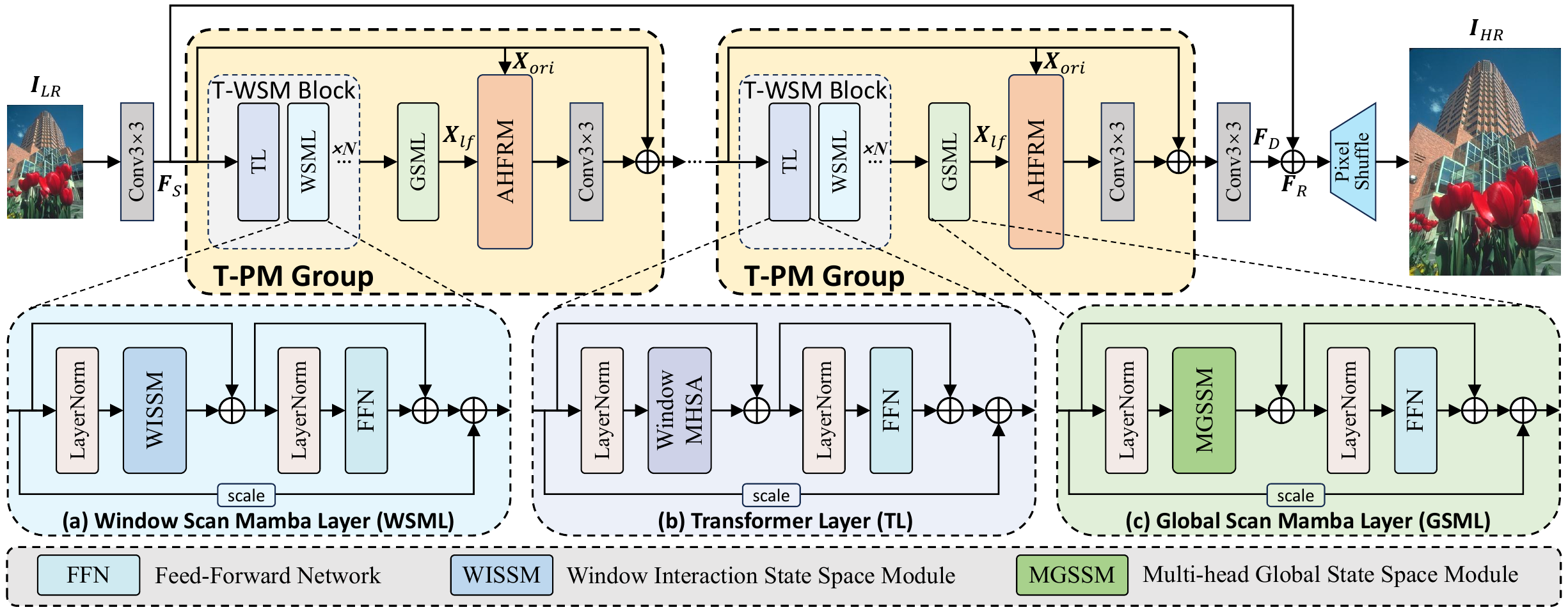} 
  \caption{The network architecture of our T-PMambaSR, as well as the framework of the (a) Window Scan Mamba Layer (WSML), (b) Transformer Layer (TL), and (c) Global Scan Mamba Layer (GSML).}
  \label{fig:pipeline}
\end{figure*}

\subsection{Frequency-based Restoration Methods}
Another line of research explicitly targets high-frequency restoration for fine textures that are often lost in spatial backbones. For instance, CRAFT~\cite{li2023feature} enhanced high-frequency features to compensate for the Transformer model's bias toward low-frequency components. ESRT~\cite{lu2022transformer} incorporated high-frequency filtering to preserve texture details, while WFEN~\cite{li2024face} reduced texture loss during downsampling by separating high-frequency features. AFFNet~\cite{huang2023adaptive} used adaptive frequency filters to dynamically mix tokens in the frequency domain, improving long-range context recovery. FourierSR~\cite{li2025fouriersr} modulated the frequency space using the Fourier frequency domain theorem to expand the receptive field of super-resolution. FDSR~\cite{xu2024fdsr} reconstructed images step-by-step using Fourier-based frequency band separation to restore high-frequency details. DMNet~\cite{li2025dual} utilized the wavelet transform to explore the relationship between high-frequency and other frequency domain features in the SR process. CFSR~\cite{wu2024transforming} improved image reconstruction by integrating gradient-based edge priors into its feed-forward network to preserve and strengthen high-frequency information. PoolNet~\cite{cui2025exploring} employed average and max pooling to implicitly separate and modulate low- and high-frequency features, using the enhanced high-frequency cues to refine spatial details. CAN~\cite{yan2025can} strengthened robustness through frequency-domain augmentation that adjusts high-frequency components while preserving key frequency details. Furthermore, while existing approaches often rely on complex frequency-domain transformations or implicit feature separation, it is crucial to recognize that the successive processing in modern Transformer and Mamba modules inherently causes high-frequency degradation due to their low-pass filtering tendencies.

To explicitly counteract this issue, rather than implicitly estimating lost details, our proposed AHFRM introduces a direct reference-guided restoration mechanism. It leverages pristine high-frequency information extracted from an unprocessed feature map to explicitly guide the restoration of attenuated high-frequency content in deeply processed features, thereby ensuring robust and sharp texture reconstruction.

\section{Proposed Method}~\label{Method}
In this section, we start by introducing the preliminaries of state space models before presenting a general architecture of our T-PMambaSR. Then, we examine the architectural details of its core components: the Window Interaction State Space Module (WISSM) and the Multi-head Global State Space Module (MGSSM). Finally, we discuss the design of our Adaptive High-Frequency Refinement Module (AHFRM).

\subsection{Preliminaries}

State Space Models (SSMs)~\cite{gu2024mamba} are a class of models that map a 1-D input sequence $x(t) \in \mathbb{R}$ to an N-D latent state $h(t) \in \mathbb{R}^N$ before projecting it to a 1-D output $y(t) \in \mathbb{R}$. Originating from classical control theory, they are particularly adept at modeling long-range dependencies. Specifically, a continuous-time SSM can be formulated as a linear Ordinary Differential Equation (ODE):
\begin{equation}
\begin{split}
    \frac{dh(t)}{dt} &= \mathbf{A}h(t) + \mathbf{B}x(t),\\
    y(t)             &= \mathbf{C}h(t) + \mathbf{D}x(t),
\end{split}
\end{equation}
where $\mathbf{A} \in \mathbb{R}^{N \times N}$ is the state transition matrix, which describes how the internal state evolves, and $\mathbf{B} \in \mathbb{R}^{N \times 1}$, $\mathbf{C} \in \mathbb{R}^{1 \times N}$, $\mathbf{D} \in \mathbb{R}$ are projection matrices that map the input to state, the state to output, and the input to output, respectively.

\begin{figure*}[t]   
  \centering
  \includegraphics[width=\linewidth]{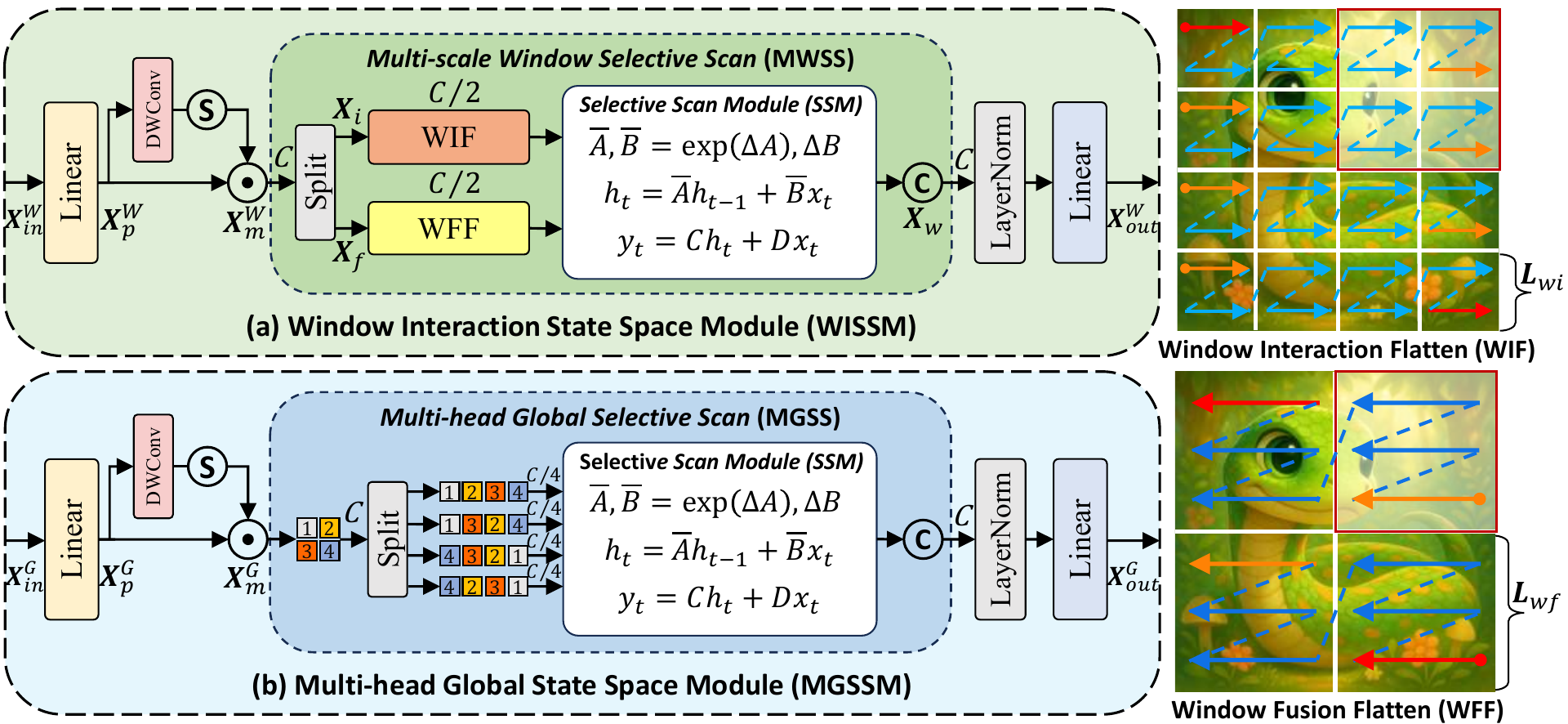} 
  \caption{The illustration of our (a) Window Interaction State Space Module (WISSM) with its two flatten mechanisms, Window Interaction / Fusion Flatten (WIF and WFF), and (b) Multi-head Global State Space Module (MGSSM).}
  \label{fig:2SSM}
\end{figure*}

To be applied to discrete sequential data, such as text or flattened image patches, this continuous system must be discretized. A common method is the zero-order hold (ZOH), which transforms the continuous parameters $(\mathbf{A}, \mathbf{B})$ into discrete counterparts $(\bar{\mathbf{A}}, \bar{\mathbf{B}})$ using a timescale parameter $\Delta$. The discretized SSM is then formulated as
\begin{equation}
\begin{split}
    h_k &= \bar{\mathbf{A}}h_{k-1} + \bar{\mathbf{B}}x_k,\\
    y_k &= \mathbf{C}h_k + \mathbf{D}x_k,
\end{split}
\end{equation}
where $\bar{\mathbf{A}} = \exp(\Delta \mathbf{A})$ and $\bar{\mathbf{B}} = (\exp(\Delta \mathbf{A}) - \mathbf{I})\mathbf{A}^{-1}\mathbf{B}$. This discrete formulation can be unrolled and computed efficiently as a global convolution, allowing it to be integrated into deep neural networks.

While the Structured State Space Sequence Model (S4) made SSMs computationally efficient by structuring the $\mathbf{A}$ matrix, its parameters remain static and input-invariant. Mamba~\cite{gu2024mamba} enhances this by introducing a selection mechanism that renders key parameters ($\mathbf{B}$, $\mathbf{C}$, $\Delta$) input-dependent. This enables Mamba to selectively modulate information flow based on content, achieving state-of-the-art performance with linear complexity. Overall, our work aims to leverage this efficient and content-aware modeling for SR.

\subsection{General Architecture}

As shown in Fig.~\ref{fig:pipeline}, given a low-resolution (LR) input image $\boldsymbol{I}_{LR} \in \mathbb{R}^{H \times W \times 3}$, an initial 3$\times$3 convolutional layer first map the input into a shallow feature $\boldsymbol{F}_{S} \in \mathbb{R}^{H \times W \times C}$, where $H$, $W$, and $C$ represent the height, width, and channel dimensions of the input image. This shallow feature $\boldsymbol{F}_{S}$ is then fed into several Transformer-Progressive Mamba (T-PM) Groups to extract deep features $\boldsymbol{F}_{D} \in \mathbb{R}^{H \times W \times C}$. Each T-PM Group consists of \(N\) stacked Transformer-Window Scan Mamba (T-WSM) Blocks that progressively refine the features. A Global Scan Mamba Layer (GSML) is then applied to expand the receptive field for global modeling, followed by an Adaptive High-Frequency Refinement Module (AHFRM) to restore high-frequency details. An additional convolutional layer is applied at the end of each group for further feature enhancement. Finally, a long skip connection combines the shallow and deep features through element-wise addition $\boldsymbol{F}_{R} = \boldsymbol{F}_{D} + \boldsymbol{F}_{S}$, and the resulting fused feature map $\boldsymbol{F}_{R}$ is upsampled to generate the HR output $\boldsymbol{I}_{HR}$.





\subsection{Window Interaction State Space Module}
As a core component of our Window Scan Mamba Layer (WSML) designed to facilitate long-range communication across windows and progressively expand the receptive field, we introduce the Window Interaction State Space Module (WISSM). As shown in Fig.~\ref{fig:2SSM} (a), WISSM first projects the input feature $\boldsymbol{X}^{W}_{in} \in \mathbb{R}^{H \times W \times C}$ with a linear layer. The projected feature is then divided into two parallel streams. The first stream acts directly as the content feature, while the second is passed through a gating mechanism to generate a spatial gate. This gate weight is then multiplied element-wise with the content feature to adaptively select and enhance important information while suppressing noise:
\begin{equation}
\begin{split}
    \boldsymbol{X}^{W}_{p} &= \operatorname{Linear}(\boldsymbol{X}^{W}_{in}), \\
    \boldsymbol{X}^{W}_{m} &= \boldsymbol{X}^{W}_{p} \odot \sigma(\operatorname{DWConv}(\boldsymbol{X}^{W}_{p})),
\end{split}
\end{equation}
where $\odot$ denotes element-wise multiplication and $\sigma$ is the sigmoid function. Subsequently, the feature $\boldsymbol{X}^{W}_{m}$ is processed by our Multi-scale Window Selective Scan (MWSS) to establish multi-scale and long-range window-based dependencies. The process can be described as:
\begin{equation}
    \boldsymbol{X}^{W}_{out} = \operatorname{Linear}(\operatorname{LN}(\operatorname{MWSS}(\boldsymbol{X}^{W}_{m}))).
\end{equation}
\textbf{Multi-scale Window Selective Scan (MWSS).} Prior works based on SSM~\cite{guo2024mambair,weng2024mamballie,zou2024wavemamba} typically employ the 2D Selective Scan for global modeling. However, in hybrid Transformer-Mamba architectures, this creates an abrupt leap from the local receptive field of window attention to the global scan, hindering a smooth multi-scale feature transition. Moreover, both window attention and the standard 2D selective scan are ill-suited for long-range interactions between discrete, window-shaped feature regions of specific size, which is vital for maintaining coherence in the restoration of global features. To this end, we propose MWSS to address these challenges.

\begin{figure*}[!htbp]
  \centering
  \includegraphics[width=\linewidth]{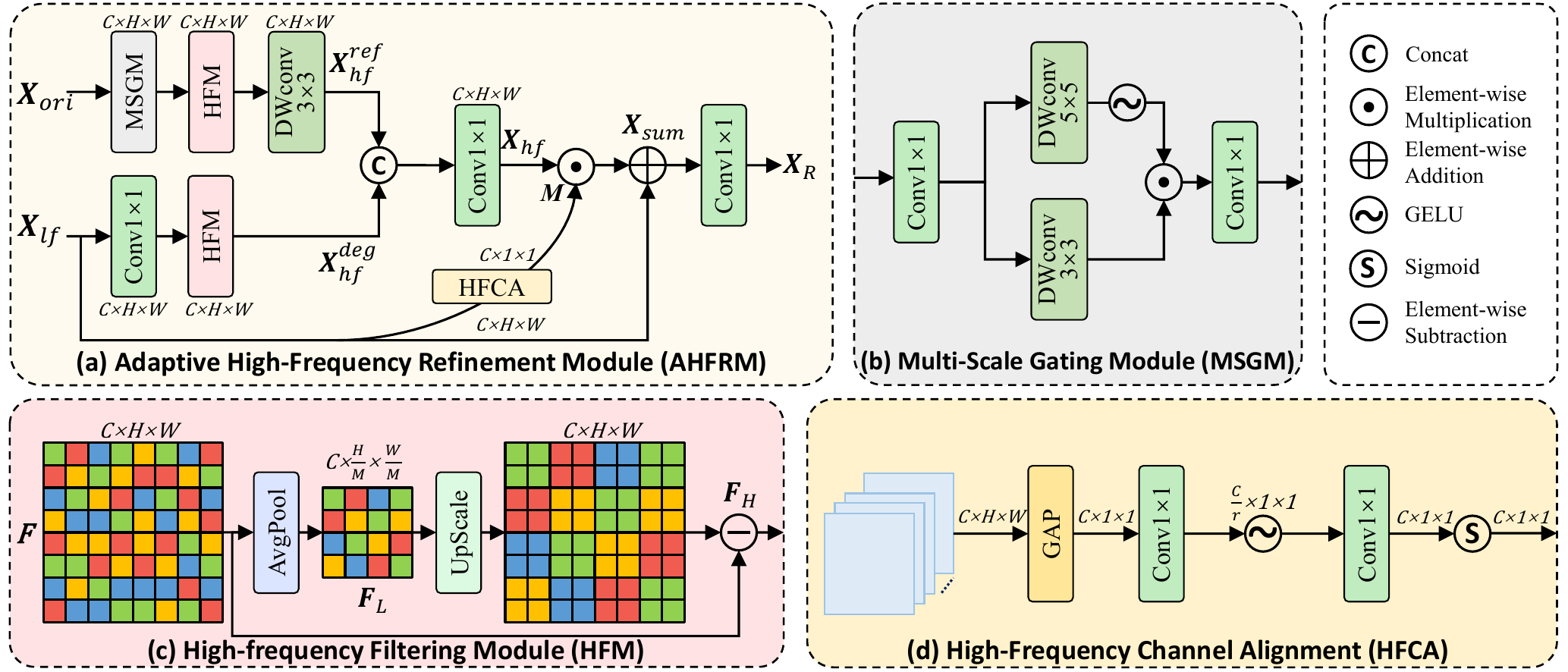}
  \caption{The architecture of our (a) Adaptive High-Frequency Refinement Module (AHFRM), (b) Multi-Scale Gating Module (MSGM), (c) High-frequency Filtering Module (HFM), and (d) High-Frequency Channel Alignment (HFCA).}
  \label{fig:AHFRM}
\end{figure*}

Differing from the traditional 2D selective scan paradigm of feature replication, multi-directional scanning, and summation, our MWSS operation enhances efficiency by splitting the feature tensor into two halves along the channel dimension. These two parts are first flattened separately by a Window Interaction Flatten (WIF) and a Window Fusion Flatten (WFF) operation. As depicted in Fig.~\ref{fig:2SSM}, WIF flattens the input feature with its scanning window size $\boldsymbol{L}_{wi}$ set to 16, consistent with the window self-attention, to facilitate long-range interactions between windows. Meanwhile, WFF sets $\boldsymbol{L}_{wf}$ to 32 to serve as a smooth transition from local to global receptive fields. Within each MWSS, WIF and WFF flatten feature maps on the same axis (\eg, both horizontal or vertical) but in opposite directions, while the flattening axes and directions alternate across different MWSS to ensure comprehensive coverage of all four raster-scan orders (\ie, top-left to bottom-right, bottom-right to top-left, top-right to bottom-left, and bottom-left to top-right). Subsequently, these two flattened feature sets are processed independently by the Selective Scan Module, after which their resulting outputs are concatenated:
\begin{equation}
\begin{aligned}
\boldsymbol{X}_{i}, \boldsymbol{X}_{f}
    &= \operatorname{Split}(\boldsymbol{X}^{W}_{m}), \\
\boldsymbol{X}_{w}
    &= \operatorname{Concat}\left(
       \begin{aligned}
       &\operatorname{SSM}(\operatorname{WIF}(\boldsymbol{X}_{i})),\\
       &\operatorname{SSM}(\operatorname{WFF}(\boldsymbol{X}_{f}))
       \end{aligned}
       \right).
\end{aligned}
\end{equation}
This parallel design achieves efficient long-range window interaction while establishing a smooth multi-scale receptive field transition.

\subsection{Multi-head Global State Space Module}
After a sequence of window self-attention and WISSM dedicated to fine-grained, multi-scale local modeling, as shown in Fig.~\ref{fig:pipeline}, we introduce the Multi-head Global State Space Module (MGSSM) within our GSML, where our purpose is to further expand the receptive field to a global scale, ensuring that our feature representation captures both fine local details and long-range structural coherence.

As depicted in Fig.~\ref{fig:2SSM} (b), our MGSSM is structurally analogous to WISSM, differing only in its utilization of the Multi-head Global Selective Scan (MGSS). The process is:
\begin{equation}
\begin{split}
    \boldsymbol{X}^{G}_{p} &= \operatorname{Linear}(\boldsymbol{X}^{G}_{in}), \\
    \boldsymbol{X}^{G}_{m} &= \boldsymbol{X}^{G}_{p} \odot \sigma(\operatorname{DWConv}(\boldsymbol{X}^{G}_{p})), \\
    \boldsymbol{X}^{G}_{out} &= \operatorname{Linear}(\operatorname{LN}(\operatorname{MGSS}(\boldsymbol{X}^{G}_{m}))),
\end{split}
\end{equation}
where $\odot$ denotes the element-wise multiplication.
\\
\textbf{Multi-head Global Selective Scan (MGSS).} 
Adopting a paradigm similar to MWSS, we process the input tensor by first splitting it into four parts along the channel dimension, each with C/4 channels. Each part is then flattened and scanned in parallel along one of the four raster-scan orders (\ie, top-left to bottom-right, bottom-right to top-left, top-right to bottom-left, and bottom-left to top-right). The four resulting output features are subsequently concatenated. This design drastically reduces computational complexity while achieving performance that is nearly identical to that of a version using the traditional 2D selective scan, as validated in Table~\ref{tab:ablation_components}.

\subsection{Adaptive High-Frequency Refinement Module}
We propose an Adaptive High-Frequency Refinement Module (AHFRM) that extracts pristine high-frequency information from an initial, unprocessed feature as a reference. This reference guides the adaptive restoration of high-frequency information that has been attenuated within the feature after undergoing low-frequency-biased modeling by a series of Transformer and Mamba operations.

\input{tabs/Tab_benchmark}

\begin{table}[t]
\tiny
\setlength\tabcolsep{2.5pt}
\centering
\vspace{0mm}
\caption{Quantitative evaluations of our method and existing lightweight SR methods on a real-world test set.}
\vspace{-0mm}
\label{tab:realsr}
\resizebox{0.47\textwidth}{!}{
\begin{tabular}{c||c|c|c|c}
\toprule
& & & & \multicolumn{1}{c}{RealSRv3~\cite{cai2019toward}} 
\\ 
\cmidrule{5-5}
    \multicolumn{1}{c||}{\multirow{-1.85}{*}{Scale}}
    & \multicolumn{1}{c|}{\multirow{-1.85}{*}{Methods}} 
    & \multicolumn{1}{c|}{\multirow{-1.85}{*}{Params$\downarrow$}}
    & \multicolumn{1}{c|}{\multirow{-1.85}{*}{FLOPs$\downarrow$}}
    & PSNR$\uparrow$/SSIM$\uparrow$/LPIPS$\downarrow$
    \\ 
    \hline
    \multirow{5}{*}{$\times 4$}  
    & SRFormer-light~\cite{zhou2023srformer}     & 873K  &62.8G   & 29.38/0.8314/0.2697     \\
    & HiT-SIR~\cite{zhang2024hit}    & 792K  &\textbf{53.8G}   & 29.34/0.8312/0.2724     \\
    & MambaIR-light~\cite{guo2024mambair}  & 924K  &84.6G   & 29.39/0.8308/0.2706     \\
    & MambaIRv2-light~\cite{guo2025mambairv2}  & 790K  &75.6G   & 29.40/\textbf{0.8325}/0.2682    \\
    & \textbf{T-PMambaSR (Ours)}   & \textbf{703K}  & 63.0G  & \textbf{29.43}/\textbf{0.8325}/\textbf{0.2658}   \\
    \bottomrule
\end{tabular}}
\end{table}

\input{img_code/Figure_benchmark}

Inspired by the design of extracting and enhancing high-frequency features in ESRT~\cite{lu2022transformer}, as shown in Fig.~\ref{fig:AHFRM} (c), for a given input $\boldsymbol{F} \in \mathbb{R}^{C \times H \times W}$, we first apply a pooling layer to reduce its spatial resolution by half, utilizing the low-frequency extraction properties of pooling to obtain its low-frequency components, $\boldsymbol{F}_{L}\in \mathbb{R}^{C \times \frac{H}{2} \times \frac{W}{2}}$. This low-frequency feature is then upsampled back to the original resolution using bilinear interpolation. Finally, the upsampled low-frequency feature is subtracted from the original feature $\boldsymbol{F}$, and the resulting difference constitutes the desired high-frequency feature $\boldsymbol{F}_{H} \in \mathbb{R}^{C \times H \times W}$. This process can be formulated as:
\begin{equation}
\begin{split}
    \boldsymbol{F}_{L} &= \operatorname{Pooling}(\boldsymbol{F}), \\
    \boldsymbol{F}_{H} &= \boldsymbol{F} - \operatorname{Upsample}(\boldsymbol{F}_{L}).
\end{split}
\end{equation}
As depicted in Fig.~\ref{fig:AHFRM} (a), each Adaptive High-Frequency Refinement Module (AHFRM) receives two inputs: the pristine, unprocessed feature $\boldsymbol{X}_{ori} \in \mathbb{R}^{C \times H \times W}$ with its high-frequency information intact, and the deeply processed, low-frequency-biased feature $\boldsymbol{X}_{lf} \in \mathbb{R}^{C \times H \times W}$. To obtain the reference high-frequency information $\boldsymbol{X}_{hf}^{ref} \in \mathbb{R}^{C \times H \times W}$, $\boldsymbol{X}_{ori}$ is first processed by our Multi-Scale Gating Module (MSGM) for selective multi-scale extraction, followed by an HFM and a 3$\times$3 depth-wise convolutional layer. In parallel, $\boldsymbol{X}_{lf}$ is passed through a 1$\times$1 convolutional layer and an HFM to extract the degraded high-frequency information $\boldsymbol{X}_{hf}^{deg} \in \mathbb{R}^{C \times H \times W}$ targeted for adaptive restoration:
\begin{equation}
\begin{split}
    \boldsymbol{X}_{hf}^{ref} &= \operatorname{DWConv_{3\times3}}(\operatorname{HFM}(\operatorname{MSGM}(\boldsymbol{X}_{ori}))), \\
    \boldsymbol{X}_{hf}^{deg} &= \operatorname{HFM}(\operatorname{Conv_{1\times1}}(\boldsymbol{X}_{lf})).
\end{split}
\end{equation}
Then, the two extracted high-frequency features are concatenated and fused by a 1$\times$1 convolution, which projects them back to the original channel dimension to form a refined high-frequency feature $\boldsymbol{X}_{hf} \in \mathbb{R}^{C \times H \times W}$. In parallel, we introduce a High-Frequency Channel Alignment (HFCA). As shown in Fig.~\ref{fig:AHFRM} (d), HFCA generates a channel attention map $\boldsymbol{M} \in \mathbb{R}^{C \times 1 \times 1}$ from the high-frequency degraded feature  $\boldsymbol{X}_{lf} \in \mathbb{R}^{C \times H \times W}$ and uses it to apply channel-wise weights to the refined high-frequency feature. Subsequently, this weighted high-frequency feature is fused with the low-frequency-biased feature via a residual connection and a 1$\times$1 convolution to generate the refined feature $\boldsymbol{X}_{R}$. This process is:
\begin{equation}
\begin{split}
    \boldsymbol{X}_{hf} &= \operatorname{Conv_{1\times1}}(\operatorname{Concat}(\boldsymbol{X}_{hf}^{ref}, \boldsymbol{X}_{hf}^{deg})),\\
    \boldsymbol{X}_{R} &= \operatorname{Conv_{1\times1}}(\boldsymbol{X}_{hf} \odot \operatorname{HFCA}(\boldsymbol{X}_{lf}) + \boldsymbol{X}_{lf}).
\end{split}
\end{equation}
Based on these designs, our AHFRM supplements the high-frequency information weakened by the Transformer or Mamba.



\subsection{Computational Complexity Analysis}
For an input feature map of size $H \times W \times C$, standard Window-MHSA (W-MHSA) with a window size of $M \times M$ exhibits a computational complexity of $\mathcal{O}(H W C^2 + H W M^2 C)$. Although this scales linearly with the overall spatial resolution $H \times W$, it remains quadratic with respect to the window size. In contrast, the selective scan component in our proposed Mamba-based modules (WISSM and MGSSM) processes features with a complexity of $\mathcal{O}(H W C N)$, where $N$ denotes the dimension of the SSM state. By adopting different scanning paths alone, MWSS and MGSS achieve efficient receptive field expansion and long-range inter-window interactions without increasing the computational cost. This design preserves strict linearity and avoids the quadratic bottleneck caused by enlarging window sizes, thereby significantly reducing the computational burden.

\input{img_code/Figure_RealSR}

\begin{figure}[t]
    \scriptsize
    \centering
    \scalebox{1}{
    \hspace{-5.5mm}
    \begin{tabular}{lc}
        \begin{adjustbox}{valign=t}
	\begin{tabular}{c}	 
        \includegraphics[width=0.2\textwidth,height=0.1045\textheight]{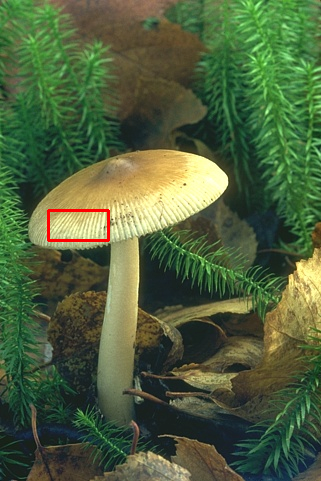} \\[-0.5mm]
		\fontsize{6.5pt}{1pt}\selectfont BSDS100 ($\times 2$): 208001 \\
	\end{tabular}
	\end{adjustbox}
	\hspace{-7mm}
    
	\begin{adjustbox}{valign=t}
	\begin{tabular}{ccc}
		\includegraphics[width=0.09\textwidth, height=0.045\textheight, frame,  cfbox=black 0.3pt -0.3pt]{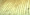} & 
		\hspace{-6.5mm}
        
        \includegraphics[width=0.09\textwidth, height=0.045\textheight, frame,  cfbox=black 0.3pt -0.3pt]{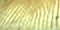} & 
		\hspace{-5mm}
		
        \includegraphics[width=0.09\textwidth, height=0.045\textheight, frame,  cfbox=black 0.3pt -0.3pt]{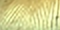} \\[-0.5mm]
  
		\fontsize{6.5pt}{1pt}\selectfont LR & \hspace{-6mm}
		\fontsize{6.5pt}{1pt}\selectfont HiT-SIR~\cite{zhang2024hit} & 
        \hspace{-5.5mm}
        \fontsize{6.5pt}{1pt}\selectfont MambaIR-light~\cite{guo2024mambair}\\			
        
        \includegraphics[width=0.09\textwidth, height=0.045\textheight, frame,  cfbox=black 0.3pt -0.3pt]{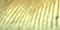} & 
		\hspace{-6.5mm}
        
        \includegraphics[width=0.09\textwidth, height=0.045\textheight, frame,  cfbox=black 0.3pt -0.3pt]{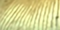} & 
		\hspace{-5mm}
		
        \includegraphics[width=0.09\textwidth, height=0.045\textheight, frame,  cfbox=black 0.3pt -0.3pt]{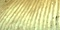} \\[-0.5mm]
		
        \fontsize{6.5pt}{1pt}\selectfont MambaIRv2-light~\cite{guo2025mambairv2} & \hspace{-6mm}
        
        \fontsize{6.5pt}{1pt}\selectfont \textbf{Ours} & 
        \hspace{-4.2mm}
		
        \fontsize{6.5pt}{1pt}\selectfont Ground Truth \\
        \end{tabular}
	\end{adjustbox}
   \end{tabular} }
   \vspace{-0mm}         
\captionsetup{format=plain,justification=raggedright,singlelinecheck=false}
   \caption{Qualitative comparisons of our T-PMambaSR with existing methods on \textbf{208001} from the BSDS100 dataset ($\times2$). Our method achieves sharper and more accurate restoration on natural textures.}
   \label{visual_mushroom}
   \vspace{-3mm}
\end{figure}

\section{Experiments}
\subsection{Experimental Settings}

\subsubsection{Datasets and Metrics} Following prior works~\cite{zhou2023srformer,zhang2024hit,li2025dual}, we train our model on the DIV2K~\cite{timofte2017ntire} dataset, which contains 800 pairs of images, and evaluate it on five benchmark datasets: Set5~\cite{bevilacqua2012low}, Set14~\cite{zeyde2010single}, BSDS100~\cite{martin2001database}, Urban100~\cite{huang2015single}, and Manga109~\cite{matsui2017sketch}. The LR images are obtained from their corresponding HR counterparts using bicubic degradation. We conduct experiments under three upscaling factors: $\times 2$, $\times 3$, and $\times 4$. To further validate our method, we train and test our model on the RealSRv3~\cite{cai2019toward} dataset. It contains real-world LR-HR image pairs obtained by adjusting the focal length of a camera on the same scene and subsequently aligning the images. To evaluate performance, we use PSNR and SSIM~\cite{wang2004image}, both calculated on the Y channel in the YCbCr space. For real-world SR tasks, we also utilize the LPIPS~\cite{zhang2018unreasonable} metric, as it aligns better with human subjective visual perception compared to traditional pixel-wise metrics.

\begin{table*}[!htbp]
\tiny
\setlength\tabcolsep{3pt}
\centering
\vspace{0mm}
\caption{Efficiency comparison with existing Transformer/Mamba-based lightweight SR methods~\cite{liang2021swinir,zhou2023srformer,li2025mair,guo2025mambairv2} on the $\times4$ scale, including the number of parameters and FLOPs, inference speed, and average PSNR/SSIM on five benchmarks. FLOPs and speed are measured on upsampled images with a spatial size of 1280$\times$720 using a single NVIDIA RTX 4090 GPU.}
\vspace{-0mm}
\label{tab:trans_mamba_performance}
\resizebox{0.96\textwidth}{!}{
\begin{tabular}{c||c|c|c|c|c|c}
    \specialrule{0.15em}{0pt}{-0.8pt} 
    \multicolumn{1}{c||}{\multirow{1.2}{*}{Methods}}
    & \multicolumn{1}{c|}{\multirow{1.2}{*}{SwinIR-light~\cite{liang2021swinir}}}
    & \multicolumn{1}{c|}{\multirow{1.2}{*}{SRFormer-light~\cite{zhou2023srformer}}}  
    & \multicolumn{1}{c|}{\multirow{1.2}{*}{MaIR-Tiny~\cite{li2025mair}}}
    & \multicolumn{1}{c|}{\multirow{1.2}{*}{MaIR-Small~\cite{li2025mair}}}
    & \multicolumn{1}{c|}{\multirow{1.2}{*}{MambaIRv2-light~\cite{guo2025mambairv2}}}
    & \multicolumn{1}{c}{\multirow{1.2}{*}{\textbf{T-PMambaSR (Ours)}}}
    \\ 
    \hline
    \multicolumn{1}{c||}{\multirow{1.1}{*}{Params$\downarrow$}}
    &  {\multirow{1.1}{*}{930K}}
    &  {\multirow{1.1}{*}{873K}}
    &  {\multirow{1.1}{*}{897K}}
    &  {\multirow{1.1}{*}{1374K}}
    &  {\multirow{1.1}{*}{\underline{790K}}}
    &  {\multirow{1.1}{*}{\textbf{703K}}}
    \\ 
    \multicolumn{1}{c||}{\multirow{1.1}{*}{FLOPs$\downarrow$}}  
    &  {\multirow{1.1}{*}{63.6G}}
    &  {\multirow{1.1}{*}{\underline{62.8G}}}
    &  {\multirow{1.1}{*}{\textbf{53.1G}}}
    &  {\multirow{1.1}{*}{136.6G}}
    &  {\multirow{1.1}{*}{75.6G}}
    &  {\multirow{1.1}{*}{63.0G}}
    \\ 
    \multicolumn{1}{c||}{\multirow{1.1}{*}{Latency$\downarrow$}}  		 
    &  {\multirow{1.1}{*}{\textbf{96ms}}}
    &  {\multirow{1.1}{*}{\underline{108ms}}}
    &  {\multirow{1.1}{*}{242ms}}
    &  {\multirow{1.1}{*}{425ms}}
    &  {\multirow{1.1}{*}{138ms}}
    &  {\multirow{1.1}{*}{124ms}}
    \\ 
    \multicolumn{1}{c||}{\multirow{1.1}{*}{PSNR$\uparrow$/SSIM$\uparrow$}}  		 
    &  {\multirow{1.1}{*}{29.26/0.8274}}
    &  {\multirow{1.1}{*}{29.38/0.8296}}
    &  {\multirow{1.1}{*}{29.35/0.8287}}
    &  {\multirow{1.1}{*}{\underline{29.47}/0.8309}}
    &  {\multirow{1.1}{*}{29.43/\underline{0.8311}}}
    &  {\multirow{1.1}{*}{\textbf{29.49}/\textbf{0.8319}}}
    \\
    \toprule
\end{tabular}}
\end{table*}

\subsubsection{Implementation Details}
Consistent with previous methods, we perform data augmentation with random horizontal flips and rotations of 90°, 180°, and 270°. Our proposed model is constructed with 4 Transformer-Progressive Mamba Groups, and within each group, we set the number of Transformer-Window Scan Mamba Blocks to $N=4$ and the channel width to 48. During training, we use a batch size of 32, and the patch size is set to $64 \times 64$. The model is trained for a total of 600K iterations using the Adam optimizer with $\beta_1=0.9$ and $\beta_2=0.99$. We optimize the network using L1 loss with a weight of 1. The initial learning rate is set to $2\times10^{-4}$ and is halved at [300K, 450K, 525K, 575K] iterations. All experiments are implemented using the PyTorch framework and conducted on two NVIDIA GeForce RTX 4090 GPUs. 

\begin{table*}[t]
\tiny
\setlength\tabcolsep{5.5pt}
\centering
\vspace{0mm}
\caption{Ablation studies of our core model components. We validate the effectiveness of our \textbf{MWSS} at the Window Scan stage (replaced in Model2 \& 3), \textbf{MGSS} at the Global Scan stage (replaced in Model1 \& 3), and \textbf{HFM} at the High-Frequency Refinement stage (removed in Model4). Our full model achieves the best performance and efficiency.}
\vspace{-0mm}
\label{tab:ablation_components}
\resizebox{0.96\textwidth}{!}{
\begin{tabular}{l||cc|cc|c|l|l|cc}
\toprule
& \multicolumn{2}{c|}{Window Scan} 
& \multicolumn{2}{c|}{Global Scan} 
& \multicolumn{1}{c|}{High-Frequency Refinement} 
& & &
\\ 
\cmidrule{2-6}
    \multicolumn{1}{l||}{\multirow{-2}{*}{Methods}} 
    & MWSS  & 2D-SS    
    & MGSS  & 2D-SS     
    & HFM
    & \multicolumn{1}{c|}{\multirow{-2}{*}{Params$\downarrow$}}
    & \multicolumn{1}{c|}{\multirow{-2}{*}{FLOPs$\downarrow$}}
    & \multicolumn{1}{c}{\multirow{-2}{*}{PSNR$\uparrow$}}
    & \multicolumn{1}{c}{\multirow{-2}{*}{SSIM$\uparrow$}}
    \\ 
    \hline
    Model1  & \Checkmark   &    &      & \Checkmark   & \Checkmark    & 706.6K  & 241.2G  & 39.16 & 0.9781  \\
    Model2  &    & \Checkmark   & \Checkmark      &    & \Checkmark    & 761.1K  & 249.8G  & 39.08  & 0.9780  \\
    Model3  &    & \Checkmark   &      & \Checkmark  & \Checkmark    & 780.3K  & 253.0G  & 39.12 & 0.9779 \\
    Model4  & \Checkmark   &   & \Checkmark     &   & \XSolidBrush    & \textbf{687.0K}  & \textbf{237.7G}  & 39.09  & 0.9781  \\
    \textbf{Ours}   & \Checkmark   &   &\Checkmark      &   & \Checkmark    & \textbf{687.0K}  & 238.1G  & \textbf{39.20} & \bf{0.9783} \\

    \bottomrule
\end{tabular}}
\end{table*}

\subsection{Comparisons with State-of-the-Art Methods}
To evaluate the performance of our model, we compare it with several state-of-the-art lightweight SR approaches, including Transformer-based methods: SwinIR-light~\cite{liang2021swinir}, ELAN-light~\cite{zhang2022efficient}, SRFormer-light~\cite{zhou2023srformer}, Omni-SR~\cite{wang2023omni}, HiT-SIR~\cite{zhang2024hit}, and ESC-lt~\cite{lee2025emulating}; Mamba-based methods: CNMC~\cite{wang2025collaborative}, MambaIR-light~\cite{guo2024mambair} and MaIR-Small~\cite{li2025mair}; as well as the hybrid Transformer-Mamba approach MambaIRv2-light~\cite{guo2025mambairv2}.

\subsubsection{Quantitative Comparisons}
Table~\ref{tab:benchmark} presents the quantitative comparison of our T-PMambaSR with existing methods at scale factors of $\times2$, $\times3$, and $\times4$. Our model outperforms existing methods with architectures based solely on Transformer or Mamba in terms of both PSNR and SSIM across most benchmarks. Compared to MambaIRv2-light~\cite{guo2025mambairv2}, a powerful hybrid Transformer-Mamba model, our method achieves comparable performance on the $\times2$ task and surpasses it on the $\times3$ and $\times4$ tasks while utilizing fewer parameters and FLOPs. Specifically, our model outperforms SRFormer-light~\cite{zhou2023srformer} on Urban100~\cite{huang2015single} by 0.26 dB, 0.25 dB, and 0.20 dB in PSNR at three scale factors, respectively, while saving 20\% of the parameters. Compared to the latest Mamba-based method, MaIR-Small~\cite{li2025mair}, our method achieves an average PSNR gain of 0.1dB on the $\times2$ task and demonstrates consistent performance advantages across all benchmarks on the scale of $\times3$ and $\times4$, while using only approximately 51\% of its parameters and 45\% of its FLOPs, showing the lightweight nature of ours.

For the real-world scenario, we conduct experiments on RealSRv3~\cite{cai2019toward}, which provides authentic LR-HR pairs captured by modulating camera focal lengths. We compare our T-PMambaSR with several lightweight models: SRFormer-light~\cite{zhou2023srformer}, HiT-SIR~\cite{zhang2024hit}, MambaIR-light~\cite{guo2024mambair}, and MambaIRv2-light~\cite{guo2025mambairv2}. For a fair comparison, all models are initialized with their pre-trained $\times4$ weights and subsequently fine-tuned for 200K iterations on the RealSRv3 training set. Evaluation is then performed on the official RealSRv3 test set. As presented in Table~\ref{tab:realsr}, our model achieves the highest PSNR, ties for the best SSIM, and obtains the lowest LPIPS among the listed methods, while using the fewest parameters.


\subsubsection{Qualitative Comparisons}
Fig.~\ref{visual_benchmark} presents the visual comparisons on synthesized datasets for the $\times2$, $\times3$, and $\times4$ tasks, while Fig.~\ref{visual_realsr} shows the $\times4$ qualitative results on the real-world test set, including Nikon and Canon datasets. Across all scales, our method demonstrates superior restoration clarity and accuracy. While other Transformer and Mamba-based models tend to blur or misalign high-frequency details, our adaptive refinement strategy excels at restoring these challenging textures, such as small letters in \textit{ppt3} and fine textures in \textit{img024} and \textit{Canon045}. Moreover, other models lacking a fine-grained receptive field transition exhibit significant distortion or indistinction on long-range structural details (\eg, \textit{img011}, \textit{img004}, and \textit{Nikon042}). In contrast, our progressive receptive field expansion enables a more accurate reconstruction of these structures, yielding results closest to the ground truth. Furthermore, to demonstrate the generality of our model beyond highly structured scenes, Fig.~\ref{visual_mushroom} visualizes its performance on natural textures using an image from the BSDS100 dataset (\textit{image\_208001}). Compared with existing models, our T-PMambaSR achieves sharper and more accurate restoration of complex natural patterns, confirming its robust generalization capability across diverse image types.

\subsubsection{Efficency}
As shown in Table~\ref{tab:trans_mamba_performance}, we evaluate the efficiency of our approach in comparison to existing Transformer- and Mamba-based methods, including SwinIR-light~\cite{liang2021swinir}, SRFormer-light~\cite{zhou2023srformer}, MaIR-Tiny~\cite{li2025mair}, MaIR-Small~\cite{li2025mair}, and MambaIRv2-light~\cite{guo2025mambairv2}. Using $\times4$ scale as an example, we report the model's parameter count, FLOPs, inference speed, and average PSNR/SSIM across five benchmarks reported in TABLE~\ref{tab:benchmark}. These results demonstrate that with the fewest parameters, our model achieves the fastest inference speed among Mamba-based methods, reducing inference latency by 49\% and 71\% compared to MaIR-Tiny and MaIR-Small, respectively, while remaining competitive with Transformer-based models.

\begin{table}[t]
\tiny
\setlength\tabcolsep{5.5pt}
\centering
\vspace{0mm}
\caption{Ablation studies of our Multi-scale Window Selective Scan (MWSS) strategy and our choice of 
window sizes in WIF and WFF, respectively. Disabling MWSS (Case1-3) or using window pairs with a large size disparity (Case4) leads to performance degradation. This confirms that a smooth multi-scale transition (Ours) is crucial.}
\vspace{-0mm}
\label{tab:ablation_window_size}
\resizebox{0.49\textwidth}{!}{
\begin{tabular}{c||c|c|cc}
\toprule
    \multicolumn{1}{c||}{\multirow{-1}{*}{Methods}} 
    & \multicolumn{1}{c|}{\multirow{-1}{*}{MWSS}} 
    & \multicolumn{1}{c|}{\multirow{-1}{*}{Window Sizes}}
    & \multicolumn{1}{c}{\multirow{-1}{*}{PSNR$\uparrow$}} 
    & \multicolumn{1}{c}{\multirow{-1}{*}{SSIM$\uparrow$}}
    \\ 
    \hline
    Case1  &  \XSolidBrush & (16$\times$16, 16$\times$16) &     39.15 &  0.9781  \\
    Case2  &  \XSolidBrush  & (32$\times$32, 32$\times$32)  &   39.13 &  0.9780     \\
    Case3  &  \XSolidBrush  & (64$\times$64, 64$\times$64)  &   39.13 &  0.9781   \\
    Case4  &  \Checkmark & (16$\times$16, 64$\times$64)  &  39.13    &    0.9779    \\
    Case5  &  \Checkmark & (32$\times$32, 64$\times$64)  &  39.16    &    0.9782    \\
    \textbf{Ours}   & \Checkmark   & (16$\times$16, 32$\times$32)   & \textbf{39.20}   & \textbf{0.9783}       \\
    \noalign{\vskip 0pt} 
    \multicolumn{5}{@{}l@{}}{\hspace{1cm}} \\
    [-2.3ex] 
    \bottomrule
\end{tabular}}
\vspace{1mm}
\end{table}

\subsection{Ablation Studies}
For ablation studies, all models are trained on the DIV2K dataset with 300K iterations and tested on the Manga109 ($\times2$) test set. Our ablation studies focus on analyzing the impact of our proposed module components, the choice of scanning window size, and the strategy for receptive field transition.

\subsubsection{Module Components}
Table~\ref{tab:ablation_components} presents our ablation study validating that our proposed MWSS, MGSS, and HFM operations each contribute positively to performance. To verify the efficacy of our scan operations, we create three variants: Model1 replaces MGSS (in GSML) with the widely used 2D-selective scan (2D-SS)~\cite{liu2024vmamba,guo2024mambair,weng2024mamballie}; Model2 replaces MWSS (in WSML) with 2D-SS; and Model3 replaces all of our scan operations (MWSS and MGSS) with 2D-SS. Concurrently, to validate the HFM's role, Model4 removes it from the AHFRM. The results demonstrate that our model, with all components, achieves the best restoration metrics with the lowest parameter count and computational cost. Notably, replacing all our custom scans (Model3) increases parameters by approximately 100K and degrades PSNR by 0.08dB. Furthermore, removing the parameter-free and computationally negligible HFM (Model4) causes a 0.11dB PSNR drop, demonstrating its significant advantage for SR reconstruction.

\subsubsection{Scanning Window Size}
To validate the contribution of our Multi-scale Window Selective Scan (MWSS) strategy and demonstrate the optimality of the selected $16 \times 16$ and $32 \times 32$ scanning window sizes, we conduct a series of ablation experiments, with results presented in Table~\ref{tab:ablation_window_size}. In Case1, Case2, and Case3, we deactivate the MWSS strategy and instead use two parallel scans with identical window sizes but different directions. In Case4 and Case5, the MWSS strategy is employed with different combinations of window sizes for the parallel scans. The results show that our proposed MWSS strategy and window size selection yield the best performance. Notably, simply adopting the MWSS strategy is insufficient if the disparity between window sizes is too large, as seen in the performance drop in Case4. We observe that a smooth receptive field transition (as in Ours) is essential for achieving optimal modeling performance.

\begin{table}[t]
\tiny
\setlength\tabcolsep{1pt}
\centering
\vspace{0mm}
\caption{Ablation studies of the receptive field transition sequence. Our progressive small-to-large strategy yields the best performance. Any deviation from this designed sequence results in performance degradation, confirming the optimality of our approach.}
\vspace{-0mm}
\label{tab:ablation_hir_receptive}
\resizebox{0.47\textwidth}{!}{
\begin{tabular}{c|c|cc}
\toprule
    \multicolumn{1}{c|}{\multirow{-1}{*}{Modeling Stage}}
    & \multicolumn{1}{c|}{\multirow{-1}{*}{Transition Sequence}}
    & \multicolumn{1}{c}{\multirow{-1}{*}{PSNR$\uparrow$}} 
    & \multicolumn{1}{c}{\multirow{-1}{*}{SSIM$\uparrow$}}
    \\ 
    \hline
    GSML$\rightarrow$TL$\rightarrow$WSML & global$\rightarrow$local$\rightarrow$regional & 39.16 & 0.9780 \\
    WSML$\rightarrow$TL$\rightarrow$GSML & regional$\rightarrow$local$\rightarrow$global & 39.13 & 0.9781 \\
    TL$\rightarrow$WSML$\rightarrow$GSML & local$\rightarrow$regional$\rightarrow$global & \textbf{39.20} & \textbf{0.9783} \\
    \noalign{\vskip 0pt} 
    \multicolumn{4}{@{}l@{}}{\hspace{1cm}} \\
    [-2.4ex] \bottomrule
\end{tabular}}
\end{table}

\input{img_code/Figure_receptive_field}

\subsubsection{Receptive Field Transition}
Inspired by prior works~\cite{zhang2024hit,guo2025mambairv2}, we design a progressive modeling strategy that expands the receptive field from small to large. We conduct an ablation study to validate the effectiveness of this sequence, with results presented in Table~\ref{tab:ablation_hir_receptive}. Our modules—TL, WSML, and GSML—are designed to model features at local, regional, and global scales, respectively. The results confirm that our proposed small-to-large (local $\rightarrow$ regional $\rightarrow$ global) sequence achieves optimal performance. As shown, any deviation from this progressive order leads to degradation in performance.

\begin{table}[t]
\tiny
\setlength\tabcolsep{2pt}
\centering
\vspace{1.5mm}
\captionsetup{type=table} 
\caption{Quantitative comparison of performance on multi-scale patches.  ``Arch'' denotes the architecture (``T'': Transformer, ``M'': Mamba, ``T\-M'': Hybrid), and ``RFT'' indicates whether a Receptive Field Transition strategy is employed. Our hybrid model with a progressive RFT strategy obtains the highest metrics across all three patch scales.}
\label{tab:receptive_field}
\resizebox{0.48\textwidth}{!}{
\begin{tabular}{c||c|c|c|c|c}
\toprule
\multirow{3}{*}{Methods} & 
\multirow{3}{*}{Arch} & 
\multirow{3}{*}{RFT} & 
\multicolumn{1}{c|}{Patch1} &
\multicolumn{1}{c|}{Patch2} &
\multicolumn{1}{c}{Patch3} \\
\cmidrule{4-6}
 & & & 
\multicolumn{1}{c|}{8$\times$16 pixels} &
\multicolumn{1}{c|}{16$\times$32 pixels} &
\multicolumn{1}{c}{32$\times$16 pixels} \\
\cmidrule{4-6}
 & & & \multicolumn{3}{c}{PSNR$\uparrow$/SSIM$\uparrow$} 
\\
\midrule
SRFormer~\cite{zhou2023srformer} & T & \XSolidBrush & 25.29/0.7997 & 19.32/0.5315  & 18.92/0.6958 \\
HiT-SIR~\cite{zhang2024hit} & T & \Checkmark & 25.25/0.7787	 & \underline{21.05}/\underline{0.7151}	 &  21.02/0.8364	\\
ESC-lt~\cite{lee2025emulating} & T & \XSolidBrush & 25.40/0.7853 & 19.48/0.5455	& 20.60/0.8233	\\
MambaIR~\cite{guo2024mambair} & M & \XSolidBrush & 24.86/0.7489 & 19.36/0.5265 & \underline{22.03}/\underline{0.8663} \\
MambaIRv2~\cite{guo2025mambairv2} & T-M & \XSolidBrush & \underline{25.92}/\underline{0.8206} & 18.93/0.4808 & 16.75/0.4775 \\
\textbf{Ours} & T-M & \Checkmark & \textbf{26.81}/\textbf{0.8719} & \textbf{21.20}/\textbf{0.7261} & \textbf{22.99}/\textbf{0.8814} \\
\bottomrule
\end{tabular}}
\end{table}

\section{Further Analysis}

\subsection{Progressive Expansion of Receptive Field}
Our method proposes a progressive receptive field expansion strategy based on a hybrid Transformer-Mamba architecture. To demonstrate the superiority of this paradigm, we conduct a qualitative comparison on the \textit{Urban100\_img073}, which is rich in multi-scale features. As shown in Fig.~\ref{visual_receptive_field}, our model achieves the best visual results and the highest PSNR/SSIM (as shown in Table~\ref{tab:receptive_field}) on objects at various scales.

Transformer-based methods~\cite{zhou2023srformer,zhang2024hit,lee2025emulating}, leveraging window-based multi-head self-attention (with a standard window size of $16 \times 16$ or smaller; here, we adopt $16 \times 16$ for our W-MHSA), excel at restoring regional features within small window sizes (\eg, \textbf{Patch 1}), where Mamba-based methods~\cite{guo2024mambair} struggle. However, when the target object's size exceeds the size of windows (\eg, \textbf{Patch 2} and \textbf{Patch 3}), these Transformer-based models show their limitations. Their fixed, small windows cannot capture the complete object, and they suffer from a lack of inter-window interaction. Meanwhile, Mamba-based models like MambaIR~\cite{guo2024mambair} and MambaIRv2~\cite{guo2025mambairv2} also fail to achieve optimal results, as their scanning mechanisms (\eg, 2D selective scan and semantic guided neighboring scan) disrupt the original spatial relationships of neighboring pixels and lack a smooth receptive field transition.

\begin{figure*}[t]
  \centering
  \includegraphics[width=\linewidth]{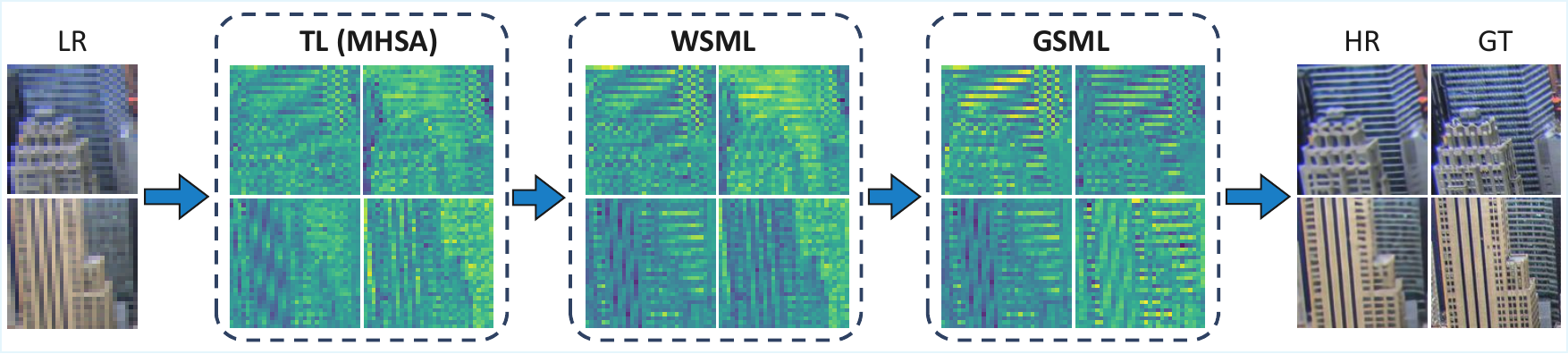}
  \caption{Visualization of features from three stages. We show the outputs from our TL and WSML (from the final T-WSM Block of the last two T-PM Groups) and the subsequent GSML. The indistinct long-range features from the TL stage are progressively detailed by the subsequent WSML (regional) and GSML (global) stages.}
  \label{fig:receptive_field}
  \vspace{-2mm}
\end{figure*}

\input{img_code/Figure_LAM}

\begin{figure*}[!htbp]
\vspace{5.5mm}
\centering
\begin{overpic}[width=0.999\linewidth]{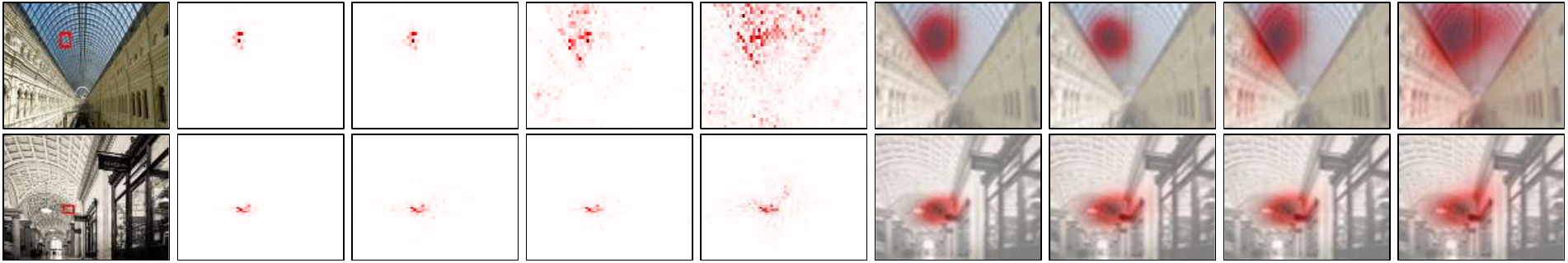}
\put(27.5,17.15){\color{black}{\fontsize{8pt}{1pt}\selectfont LAM Attribution}}
\put(70.5,17.15){\color{black}{\fontsize{8pt}{1pt}\selectfont Pixel Contribution Area}}

\put(2.2,-1.2){\color{black}{\fontsize{6.5pt}{1pt}\selectfont Ground Truth}}
\put(13.5,-1.2){\color{black}{\fontsize{6.5pt}{1pt}\selectfont Pure Mamba}}
\put(25,-1.2){\color{black}{\fontsize{6.5pt}{1pt}\selectfont Sequence1}}
\put(36,-1.2){\color{black}{\fontsize{6.5pt}{1pt}\selectfont Sequence2}}
\put(48.6,-1.2){\color{black}{\fontsize{6.5pt}{1pt}\selectfont \bf{Ours}}}

\put(58.0,-1.2){\color{black}{\fontsize{6.5pt}{1pt}\selectfont Pure Mamba}}
\put(69.5,-1.2){\color{black}{\fontsize{6.5pt}{1pt}\selectfont Sequence1}}
\put(80.5,-1.2){\color{black}{\fontsize{6.5pt}{1pt}\selectfont Sequence2}}
\put(93.1,-1.2){\color{black}{\fontsize{6.5pt}{1pt}\selectfont \bf{Ours}}}

\put(11.8,0.6){\color{black}{\fontsize{6.5pt}{1pt}\selectfont DI: 6.451}}
\put(23.0,0.6){\color{black}{\fontsize{6.5pt}{1pt}\selectfont DI: 13.012}}
\put(34.15,0.6){\color{black}{\fontsize{6.5pt}{1pt}\selectfont DI: 11.526}}
\put(45.3,0.6){\color{black}{\fontsize{6.5pt}{1pt}\selectfont \bf{DI: 19.380}}}

\put(11.8,9.0){\color{black}{\fontsize{6.5pt}{1pt}\selectfont DI: 12.065}}
\put(23.0,9.0){\color{black}{\fontsize{6.5pt}{1pt}\selectfont DI: 9.518}}
\put(34.15,9.0){\color{black}{\fontsize{6.5pt}{1pt}\selectfont DI:  17.173}}
\put(45.3,9.0){\color{black}{\fontsize{6.5pt}{1pt}\selectfont \bf{DI: 29.059}}}

\end{overpic}
\vspace{-2mm}
\captionsetup{format=plain,justification=raggedright,singlelinecheck=false}
   \caption{LAM~\cite{gu2021interpreting} comparisons at the $\times4$ scale between our progressive sequence (TL $\rightarrow$ WSML $\rightarrow$ GSML) and its variants: (1) shuffled variants (Sequence 1: WSML $\rightarrow$ TL $\rightarrow$ GSML, Sequence 2: GSML $\rightarrow$ TL $\rightarrow$ WSML), and (2) a pure Mamba variant, which replaces all W-MHSA modules with MWSS. The proposed progressive sequence yields the highest DI value and the largest pixel contribution area, confirming that our local-to-global transition architecture more effectively enables broad contextual engagement than both the pure Mamba variant and the shuffled sequences.}
\label{fig:LAM_sequence}
\vspace{-2mm}
\end{figure*}

In contrast, our method performs well in these scenarios. For larger structures (\eg, \textbf{Patch 2} and \textbf{Patch 3}), after the initial MHSA modeling, it employs the WSML as a transition. The WSML refines the preceding MHSA features via inter-window interaction and captures the complete object using larger-window modeling, before the GSML embeds the target in a global context for final refinement. This rich, progressive modeling process fully leverages the respective advantages of the hybrid Transformer-Mamba architecture. Notably, HiT-SIR~\cite{zhang2024hit} similarly recognizes the importance of progressive expansion, yet their gradually enlarging attention window only achieved suboptimal results. This is because HiT-SIR~\cite{zhang2024hit} is confined to information exchange within different windows, whereas our approach utilizes a scanning mechanism to concatenate information across different windows, thereby achieving superior reconstruction.

Fig.~\ref{fig:receptive_field} visualizes the intermediate feature maps from our three modeling stages. Details that remain indistinct within the smaller receptive field of an earlier stage are progressively refined as they pass through subsequent stages with larger contextual windows, ultimately leading to a result that is clear at all scales. Furthermore, we employ Local Attribution Maps (LAM)~\cite{gu2021interpreting} in Fig.~\ref{fig:LAM} to analyze the contextual range of different architectures. LAM illustrates the model's contextual engagement by correlating all pixels with a target patch (red), where a higher Diffusion Index (DI) or larger contribution area signifies a broader receptive field. Our method achieves the highest DI value and the largest pixel contribution area, demonstrating its ability to leverage a wider context. Although MambaIR-light~\cite{guo2024mambair} shows a comparable contribution area, its lack of a fine-grained receptive field transition and rich modeling strategies results in significantly inferior performance. This stark contrast highlights the conceptual advancement of our architecture, which achieves a strictly fine-grained, progressive receptive field transition (local $\rightarrow$ regional $\rightarrow$ global) together with effective inter-window communication. To further validate the necessity of this design, we compare our modeling sequence (TL $\rightarrow$ WSML $\rightarrow$ GSML) with shuffled variants. As shown in Fig.~\ref{fig:LAM_sequence}, our proposed configuration achieves the highest DI value and the largest pixel contribution area, significantly outperforming both the shuffled variants (\textbf{Sequence 1}: WSML $\rightarrow$ TL $\rightarrow$ GSML, and \textbf{Sequence 2}: GSML $\rightarrow$ TL $\rightarrow$ WSML) and the \textbf{Pure Mamba variant}, which replaces all W-MHSA modules with MWSS. Notably, the clear superiority over the Pure Mamba variant further confirms that the observed gains do not stem merely from Mamba's inherent long-range modeling capability, but from the architectural synergy brought by our local-to-global transition design. These results demonstrate that our designed sequence effectively maximizes contextual engagement and validates the structural rationale behind our progressive architectural design.


\subsection{High-Frequency Restoration}

Our AHFRM aims to refine the high-frequency information that is often lost after being processed by the Transformer and Mamba blocks. While existing frequency-aware methods like ESRT~\cite{lu2022transformer} and CRAFT~\cite{li2023feature} attempt to enhance high-frequency details either prior to deep spatial modeling or directly from features that have already been degraded by low-pass-like spatial blocks, our AHFRM overcomes this limitation by employing a \textbf{reference-guided restoration mechanism}. Specifically, its dual-branch design extracts \textit{pristine, undecayed} high-frequency priors from early shallow features to serve as explicit guidance. This enables highly targeted recovery by using pristine priors to adaptively restore damaged high-frequency features, rather than merely refining already degraded information. To empirically validate the effectiveness of this design, we replace our AHFRM with equivalent baseline modules—namely, the HPB from ESRT and the HFERB from CRAFT (denoted as \textbf{ESRT-variant} and \textbf{CRAFT-variant} in Table~\ref{tab:ablation_HF}, respectively). As shown, our AHFRM consistently outperforms both variants. These results provide strong empirical evidence that using pristine high-frequency priors to explicitly guide the restoration process is fundamentally more effective for accurate texture recovery than simply refining already degraded features.

\begin{figure}[t]
  \centering
  \includegraphics[width=\linewidth]{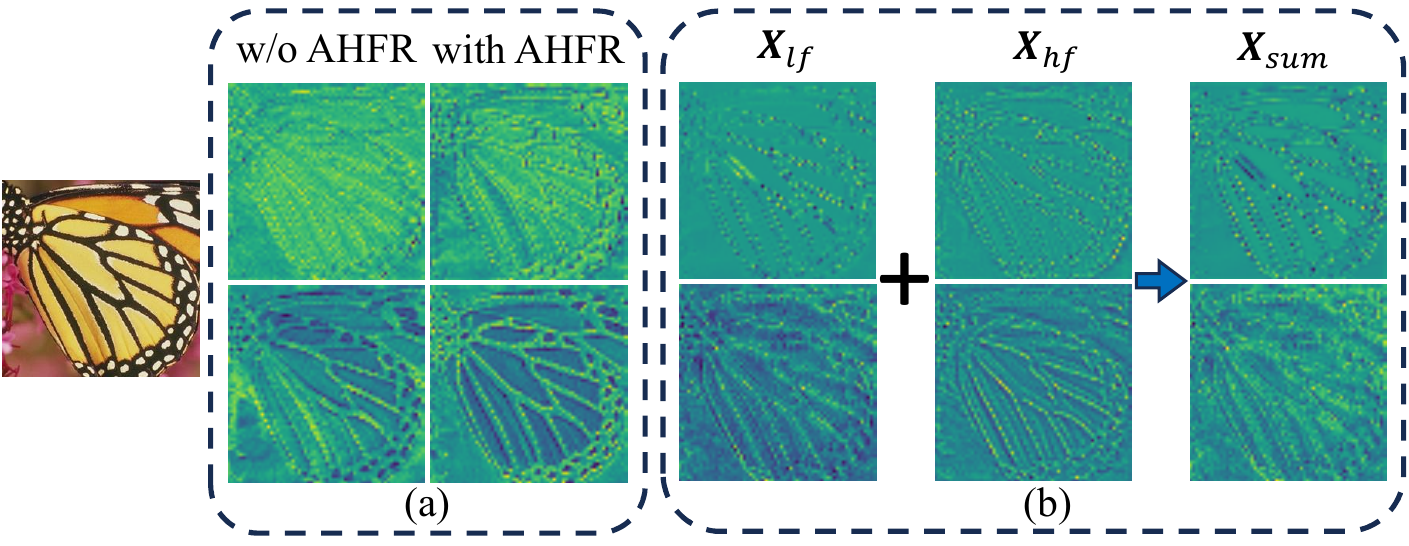}
  \caption{Visualization of our AHFRM's high-frequency refinement mechanism. (a) Demonstrates the effectiveness of AHFRM in enhancing high-frequency features. (b) It details the internal mechanism: input features are enriched by the fusion of supplementary high-frequency features, effectively restoring details missing from the input.}
  \label{fig:high_frequency}
  \vspace{-3mm}
\end{figure}

Fig.~\ref{fig:high_frequency} visualizes the effectiveness of our AHFRM using its intermediate features. Fig.~\ref{fig:high_frequency} (a) shows the features before and after the AHFRM module; the attenuated high-frequency textures become significantly clearer after modeling because we use undecayed high-frequency information as a supplement to guide the restoration of the degraded HF content. Fig.~\ref{fig:high_frequency} (b) further demonstrates the superiority of this method. The restored high-frequency feature $\boldsymbol{X}_{hf}$ in Fig.~\ref{fig:AHFRM} possesses visibly clearer textures than the degraded feature $\boldsymbol{X}_{lf}$. After an adaptive and weighted fusion, the resulting feature $\boldsymbol{X}_{sum}$ has shown superior, clearer edges prior to further convolutional processing, which confirms that our high-frequency targeted refinement is significantly effective.


\begin{table}[t]
    \centering
    \renewcommand{\arraystretch}{1.2}
    \caption{Ablation studies of our AHFRM compared with existing modules from ESRT~\cite{lu2022transformer} and CRAFT~\cite{li2023feature}. 
    }
    \label{tab:ablation_HF}
    \resizebox{0.47\textwidth}{!}{
    \hspace{-3mm}
    \begin{tabular}{@{}l||c|c|c|cc@{}}
    \toprule
        \multirow{2}{*}{Methods} & \multirow{2}{*}{Scale} & \multirow{2}{*}{Params$\downarrow$} & \multirow{2}{*}{FLOPs$\downarrow$}
        & \multicolumn{2}{c}{{Manga109~\cite{matsui2017sketch}}} \\
        \cmidrule{5-6}
        & & & & {PSNR$\uparrow$} & {SSIM$\uparrow$} \\
        \midrule
        \raisebox{-0.2mm}{ESRT-variant}
        &\raisebox{-0.2mm}{$\times4$} 
        &\raisebox{-0.2mm}{702K} 
        &\raisebox{-0.2mm}{62.6G} 
        &\raisebox{-0.2mm}{30.65} 
        &\raisebox{-0.2mm}{0.9109} \\
        \raisebox{-0.2mm}{CRAFT-variant}
        &\raisebox{-0.2mm}{$\times4$} 
        &\raisebox{-0.2mm}{670K} 
        &\raisebox{-0.2mm}{61.0G} 
        &\raisebox{-0.2mm}{30.66} 
        &\raisebox{-0.2mm}{0.9104} \\
        \raisebox{-0mm}{\textbf{Ours}}
        &\raisebox{-0mm}{$\times4$} 
        &\raisebox{-0mm}{703K} 
        &\raisebox{-0mm}{63.0G} 
        &\raisebox{-0mm}{\textbf{30.74}} 
        &\raisebox{-0mm}{\textbf{0.9110}} \\
        \bottomrule
    \end{tabular}%
    }
    \vspace{0mm}
\end{table}

\subsection{Superiority of the Hybrid Architecture}

Our T-PMambaSR deliberately integrates Transformer and Mamba modules to leverage their complementary strengths. While W-MHSA excels at extracting precise local features, expanding its receptive field (window size) incurs quadratic complexity, forcing pure-attention methods to rely on lossy operations that compromise spatial integrity. In contrast, purely Mamba-based models efficiently capture global contexts but often struggle to accurately reconstruct fine-grained local structures. Our hybrid architecture addresses these limitations. By transitioning from W-MHSA to our Mamba-based modules (MWSS and MGSS), we expand the receptive field and achieve effective inter-window communication solely by altering the selective scan paths. This ensures a smooth, fine-grained scale transition without additional costs or spatial feature loss.

To quantitatively validate our hybrid design, we construct two single-architecture baselines under a controlled parameter count: a \textbf{Pure Transformer} variant (using only hierarchical transformer blocks from HiT-SIR~\cite{zhang2024hit}) and a \textbf{Pure Mamba} variant (replacing all W-MHSA with MWSS). In these comparisons, AHFRM is kept unchanged to strictly isolate the effect of the spatial backbone. As shown in Table~\ref{tab:ablation_hybrid}, our hybrid architecture significantly outperforms both variants. These results demonstrate that integrating Transformer and Mamba modules achieves a superior receptive field transition and better overall performance than either single paradigm.

\begin{table}[t]
    \centering
    \renewcommand{\arraystretch}{1.2}
    \caption{Ablation studies comparing our hybrid architecture with pure-Transformer and pure-Mamba variants. 
    }
    \label{tab:ablation_hybrid}
    \resizebox{0.47\textwidth}{!}{
        \hspace{-3mm}
    \begin{tabular}{@{}l||c|c|c|cc@{}}
    \toprule
        \multirow{2}{*}{Methods} & \multirow{2}{*}{Scale} & \multirow{2}{*}{Params$\downarrow$} & \multirow{2}{*}{FLOPs$\downarrow$}
        & \multicolumn{2}{c}{{Manga109~\cite{matsui2017sketch}}} \\
        \cmidrule{5-6}
        & & & & \raisebox{-0.1ex}{PSNR$\uparrow$} & \raisebox{-0.1ex}{SSIM$\uparrow$} \\
        \midrule
        \raisebox{-0.2mm}{Pure-Transformer}
        &\raisebox{-0.2mm}{$\times4$} 
        &\raisebox{-0.2mm}{737K} 
        &\raisebox{-0.2mm}{53.5G} 
        &\raisebox{-0.2mm}{30.57} 
        &\raisebox{-0.2mm}{0.9087} \\
        \raisebox{-0.2mm}{Pure-Mamba}
        &\raisebox{-0.2mm}{$\times4$} 
        &\raisebox{-0.2mm}{690K} 
        &\raisebox{-0.2mm}{51.4G} 
        &\raisebox{-0.2mm}{30.36} 
        &\raisebox{-0.2mm}{0.9061} \\
        \raisebox{-0.2mm}{\textbf{Ours}}
        &\raisebox{-0.2mm}{$\times4$} 
        &\raisebox{-0.2mm}{703K} 
        &\raisebox{-0.2mm}{63.0G} 
        &\raisebox{-0.2mm}{\textbf{30.74}} 
        &\raisebox{-0.2mm}{\textbf{0.9110}} \\[0.4ex]
        \bottomrule
    \end{tabular}%
    }
\end{table}

\section{Limitations and Future Works}
While our T-PMambaSR achieves competitive performance with high efficiency, we acknowledge several limitations that offer avenues for future research. First, although our hybrid design is lightweight, the Window-MHSA component still accounts for a notable portion of the model's parameters and computational load. Exploring lighter-weight attention variants or dynamic attention mechanisms could further reduce the computational overhead without compromising their local modeling capabilities. Second, our progressive scanning strategy relies on pre-defined, fixed window sizes (\eg, $16 \times 16$ and $32 \times 32$) for the WSML. While effective, this fixed-scale approach may not be optimal for all image content. A future direction would be to develop a more flexible or content-adaptive scanning mechanism, where the window sizes and scanning paths are dynamically determined based on the input image's structural properties. Finally, our model is specifically designed and optimized for the task of single-image SR. Its effectiveness and generalizability to other low-level vision tasks~\cite{peng2025pixel}, such as image denoising and deblurring, have not yet been validated. Adapting and evaluating our progressive hybrid modeling paradigm for these related tasks remains a promising area for future investigation.

\section{Conclusion}~\label{Conclusion}
In this paper, we proposed the T-PMambaSR, an efficient hybrid Transformer-Mamba architecture for lightweight SR. Our core contribution is a progressive receptive field expansion strategy that synergizes the local modeling strength of Transformers with Mamba's long-range dependency capabilities. This is achieved through two novel modules: the WSML for regional modeling and the GSML for global context. Furthermore, we design an AHFRM to effectively restore high-frequency texture details that may be lost during global interaction. Experiments demonstrate that our T-PMambaSR achieves competitive performance over existing lightweight Transformer- and Mamba-based SR methods on both synthetic and real-world benchmarks, while incurring lower computational cost and faster inference speed. Overall, we can conclude that our work validates the superiority of a smooth, progressive local-to-global paradigm for hybrid SR models.

\bibliographystyle{IEEEtran}
\bibliography{sample-base}


\begin{IEEEbiographynophoto}{Sichen Guo}
is currently pursuing the B.E. degree in Computer Science with the Bell Honors School, Nanjing University of Posts and Telecommunications, Nanjing, China. He will pursue the M.S. degree in computer vision with the Robotics Institute, School of Computer Science, Carnegie Mellon University. His research interests include image restoration and multimodal learning.
\end{IEEEbiographynophoto}

\begin{IEEEbiographynophoto}{Wenjie Li}
received the M.S. degree in control science and engineering from the College of Automation, Nanjing University of Posts and Telecommunications, Nanjing, in 2023. He is currently pursuing the Ph.D. degree in artificial intelligence with the School of Artificial Intelligence, Beijing University of Posts and Telecommunications. His research interests include image restoration.
\end{IEEEbiographynophoto}

\begin{IEEEbiographynophoto}{Yuanyang Liu}
is currently pursuing the B.E. degree in communication engineering with the Bell Honors School, Nanjing University of Posts and Telecommunications, Nanjing, China. He will pursue the M.S. degree with the School of Artificial Intelligence, Nanjing University of Posts and Telecommunications. His research interests include 3D point cloud semantic segmentation and image restoration.
\end{IEEEbiographynophoto}

\begin{IEEEbiographynophoto}{Guangwei Gao}
(Senior Member, IEEE) received the Ph.D. degree in Pattern Recognition and Intelligent Systems from the Nanjing University of Science and Technology (NJUST), Nanjing, China, in 2014. He is currently a Professor with the School of Computer Science and Engineering, NJUST. He has served as an Associate Editor for \textsc{Pattern Recognition} and \textsc{IEEE Transactions on Image Processing}. His research interests include pattern recognition and computer vision. Personal website: \textit{https://guangweigao.github.io}.
\end{IEEEbiographynophoto}

\begin{IEEEbiographynophoto}{Jian Yang}
(Member, IEEE) received the Ph.D. degree in Pattern Recognition and Intelligent Systems from the Nanjing University of Science and Technology (NJUST), Nanjing, China, in 2002. 
He is currently a professor with the School of Computer Science and Engineering, NJUST. He is the author of more than 400 scientific papers in pattern recognition and computer vision. 
His research interests include pattern recognition, computer vision, and machine learning. He is/was an associate editor for \textsc{Pattern Recognition} and \textsc{IEEE Transactions on Neural Networks and Learning Systems}. He is a fellow of IAPR.
\end{IEEEbiographynophoto}

\begin{IEEEbiographynophoto}{Chia-Wen Lin}
(Fellow, IEEE) received the Ph.D. degree in Electrical Engineering from National Tsing Hua University (NTHU), Hsinchu, Taiwan, in 2000. 
He is currently a Distinguished Professor with the Department of Electrical Engineering and the Institute of Communications Engineering, NTHU. 
His research interests include image and video processing, computer vision, and video networking. Currently, he is serving as an Associate Editor-in-Chief of \textsc{IEEE Transactions on Circuits and Systems for Video Technology}.
\end{IEEEbiographynophoto}


\end{document}

%% file: tabs/Tab_benchmark.tex
\begin{table*}[!ht]
    \centering
    \renewcommand{\arraystretch}{1.3}
    \caption{Quantitative evaluation of lightweight SR methods, including Transformer-based and Mamba-based methods. FLOPs are measured on upsampled images of size $1280\times720$.
    The best and second-best are \textbf{bold} and \underline{underlined}.}
    \label{tab:benchmark}
    \resizebox{\textwidth}{!}{
    \belowrulesep=0pt
    \aboverulesep=-2pt
    \begin{tabular}{@{}l|c|c|c|cc|cc|cc|cc|ccc@{}}
       \specialrule{0.08em}{0pt}{0pt} 
        \multirow{2}{*}{Methods} & \multirow{2}{*}{Scale} & \multirow{2}{*}{Params$\downarrow$} & \multirow{2}{*}{FLOPs$\downarrow$}
        & \multicolumn{2}{c|}{Set5~\cite{bevilacqua2012low}}
        & \multicolumn{2}{c|}{Set14~\cite{zeyde2010single}}
        & \multicolumn{2}{c|}{BSDS100~\cite{martin2001database}}
        & \multicolumn{2}{c|}{Urban100~\cite{huang2015single}}
        & \multicolumn{2}{c}{Manga109~\cite{matsui2017sketch}} \\
        \cmidrule{5-14}
        & & & & PSNR$\uparrow$ & SSIM$\uparrow$ & PSNR$\uparrow$ & SSIM$\uparrow$ & PSNR$\uparrow$ & SSIM$\uparrow$ & PSNR$\uparrow$ & SSIM$\uparrow$ & PSNR$\uparrow$ & SSIM$\uparrow$ \\
        \midrule
        \midrule
        SwinIR-light~\cite{liang2021swinir}   & $\times2$ & 910K & 244.4G & 38.14 & 0.9611 & 33.86 & 0.9206 & 32.31 & 0.9012 & 32.76 & 0.9340 & 39.12 & 0.9783 \\
        ELAN-light~\cite{zhang2022efficient}  & $\times2$ & 621K & 201.3G & 38.17 & 0.9611 & 33.94 & 0.9207 & 32.30 & 0.9012 & 32.76 & 0.9340 & 39.11 & 0.9782 \\
        SRFormer-light~\cite{zhou2023srformer}& $\times2$ & 853K & 236.2G & 38.23 & 0.9613 & 33.94 & 0.9209 & \textbf{32.36} & \underline{0.9019} & 32.91 & 0.9353 & 39.28 & \underline{0.9785} \\
        Omni-SR~\cite{wang2023omni}           & $\times2$ & 772K & 194.5G & 38.22 & 0.9613 & \underline{33.98} & 0.9210 & \textbf{32.36} & \textbf{0.9020} & 33.05 & 0.9363 & 39.28 & 0.9784 \\
        HiT-SIR~\cite{zhang2024hit}           & $\times2$ & 772K & 209.9G & 38.22 & 0.9613 & 33.91 & 0.9213 & \underline{32.35} & \underline{0.9019} & 33.02 & 0.9365 & \textbf{39.38} & 0.9782 \\
        ESC-lt~\cite{lee2025emulating}        & $\times2$ & 603K & 359.4G & \underline{38.24} & \underline{0.9615} & \underline{33.98} & 0.9211 & \underline{32.35} & \textbf{0.9020} & 33.05 & 0.9363 & 39.33 & \textbf{0.9786} \\
        \hdashline
        CNMC~\cite{wang2025collaborative}     & $\times2$ & 769K & —
        & 38.14 & 0.9610 & 33.84 & 0.9203 & 32.29 & 0.9011 & 32.74 & 0.9337 & 39.13 & 0.9775 \\
        MambaIR-light~\cite{guo2024mambair}   & $\times2$ & 905K & 334.2G & 38.13 & 0.9610 & 33.95 & 0.9208 & 32.31 & 0.9013 & 32.85 & 0.9349 & 39.20 & 0.9782 \\
        MaIR-Small~\cite{li2025mair}          & $\times2$ & 1,355K & 542.0G   & 38.20 & 0.9611 & 33.91 & 0.9209 & 32.34 & 0.9016 & 32.97 & 0.9359 & 39.32 & 0.9779 \\
        MambaIRv2-light~\cite{guo2025mambairv2}&$\times2$ & 774K & 286.3G & \textbf{38.26} & \underline{0.9615} & \textbf{34.09} & \textbf{0.9221} & \textbf{32.36} & \underline{0.9019} & \textbf{33.26} & \textbf{0.9378} & 39.35 & \underline{0.9785} \\
        \textbf{T-PMambaSR (ours)}            & $\times2$ & 687K & 238.1G & \textbf{38.26} & \textbf{0.9616} & \textbf{34.09} & \underline{0.9219} & \textbf{32.36} & \underline{0.9019} & \underline{33.17} & \underline{0.9371} & \underline{39.36} & 0.9782 \\[0.4ex]
        \midrule
        \midrule
        SwinIR-light~\cite{liang2021swinir}   & $\times3$ & 918K & 110.8G & 34.62 & 0.9289 & 30.54 & 0.8463 & 29.20 & 0.8082 & 28.66 & 0.8624 & 33.98 & 0.9478 \\
        ELAN-light~\cite{zhang2022efficient}  & $\times3$ & 629K & 89.5G  & 34.61 & 0.9288 & 30.55 & 0.8463 & 29.21 & 0.8081 & 28.69 & 0.8624 & 34.00 & 0.9478 \\
        SRFormer-light~\cite{zhou2023srformer}& $\times3$ & 861K & 104.8G & 34.67 & 0.9296 & 30.57 & 0.8469 & 29.26 & 0.8099 & 28.81 & 0.8655 & 34.19 & 0.9489 \\
        Omni-SR~\cite{wang2023omni}           & $\times3$ & 780K &  88.4G & 34.70 & 0.9294 & 30.57 & 0.8469 & \underline{29.28} & 0.8094 & 28.84 & 0.8656 & 34.22 & 0.9487 \\
        HiT-SIR~\cite{zhang2024hit}           & $\times3$ & 780K & 94.2G  & 34.72 & 0.9298 & 30.62 & 0.8474 & 29.27 & 0.8101 & 28.93 & 0.8673 & 34.40 & 0.9496 \\
        ESC-lt~\cite{lee2025emulating}        & $\times3$ & 612K & 162.8G & 34.61 & 0.9295 & 30.52 & 0.8475 & 29.26 & 0.8102 & 28.93 & 0.8679 & 34.33 & 0.9495 \\
        \hdashline
        CNMC~\cite{wang2025collaborative}      & $\times3$ & 775K & —
        & 34.57 & 0.9285 & 30.52 & 0.8456 & 29.19 & 0.8080 & 28.61 & 0.8614 & 34.02 & 0.9471 \\
        MambaIR-light~\cite{guo2024mambair}   & $\times3$ & 913K & 148.5G & 34.63 & 0.9288 & 30.54 & 0.8459 & 29.23 & 0.8084 & 28.70 & 0.8631 & 34.12 & 0.9479 \\
        MaIR-Small~\cite{li2025mair}         & $\times3$ & 1,363K& 241.4G & \underline{34.75} & \underline{0.9300} & 30.63 & 0.8479 & \textbf{29.29} & \underline{0.8103} & 28.92 & 0.8676 & \textbf{34.46} & \underline{0.9497} \\
        MambaIRv2-light~\cite{guo2025mambairv2}&$\times3$ & 781K & 126.7G & 34.71 & 0.9298 & \textbf{30.68} & \underline{0.8483} & 29.26 & 0.8098 & \underline{29.01} & \underline{0.8689} & 34.41 & \underline{0.9497} \\
        \textbf{T-PMambaSR (ours)}            & $\times3$ & 694K & 105.6G & \textbf{34.80} & \textbf{0.9304} & \underline{30.65} & \textbf{0.8485} & \textbf{29.29} & \textbf{0.8104} & \textbf{29.06} & \textbf{0.8699} & \underline{34.45} & \textbf{0.9500} \\[0.4ex]
        \midrule 
        \midrule
        SwinIR-light~\cite{liang2021swinir}   & $\times4$ & 930K & 63.6G  & 32.44 & 0.8976 & 28.77 & 0.7858 & 27.69 & 0.7406 & 26.47 & 0.7980 & 30.92 & 0.9151 \\
        ELAN-light~\cite{zhang2022efficient}  & $\times4$ & 640K & 53.7G  & 32.43 & 0.8975 & 28.78 & 0.7858 & 27.69 & 0.7406 & 26.54 & 0.7982 & 30.92 & 0.9150 \\
        SRFormer-light~\cite{zhou2023srformer}& $\times4$ & 873K & 62.8G  & 32.51 & 0.8988 & 28.82 & 0.7872 & 27.73 & 0.7422 & 26.67 & 0.8032 & 31.17 & 0.9165 \\
        Omni-SR~\cite{wang2023omni}           & $\times4$ & 792K & 50.9G  & 32.49 & 0.8988 & 28.78 & 0.7859 & 27.71 & 0.7415 & 26.64 & 0.8018 & 31.02 & 0.9151 \\
        HiT-SIR~\cite{zhang2024hit}           & $\times4$ & 792K & 53.8G  & 32.51 & 0.8991 & 28.84 & 0.7873 & 27.73 & 0.7424 & 26.71 & 0.8045 & 31.23 & 0.9176 \\
        ESC-lt~\cite{lee2025emulating}        & $\times4$ & 624K & 91.0G    & 32.52 & 0.8995 & \underline{28.87} & 0.7878 & 27.72 & 0.7423 & 26.76 & 0.8058 & 31.26 & 0.9173 \\
        \hdashline
        CNMC~\cite{wang2025collaborative}     & $\times4$ & 782K & —
        & 32.39 & 0.8975 & 28.73 & 0.7851 & 27.68 & 0.7402 & 26.47 & 0.7978 & 30.96 & 0.9135 \\
        MambaIR-light~\cite{guo2024mambair}   & $\times4$ & 924K & 84.6G  & 32.42 & 0.8977 & 28.74 & 0.7847 & 27.68 & 0.7400 & 26.52 & 0.7983 & 30.94 & 0.9135 \\
        MaIR-Small~\cite{li2025mair}          & $\times4$ & 1,374K & 136.6G   & \textbf{32.62} & \underline{0.8998} & \textbf{28.90} & \underline{0.7882} & \textbf{27.77} & \underline{0.7431} & 26.73 & 0.8049 & \textbf{31.34} & \underline{0.9183} \\
        MambaIRv2-light~\cite{guo2025mambairv2}&$\times4$ & 790K & 75.6G  & 32.51 & 0.8992 & 28.84 & 0.7878 & \underline{27.75} & 0.7426 & \underline{26.82} & \underline{0.8079} & 31.24 & 0.9182 \\
        \textbf{T-PMambaSR (ours)}            & $\times4$ & 703K & 63.0G & \underline{32.61} & \textbf{0.9000} & \textbf{28.90} & \textbf{0.7887} & \textbf{27.77} & \textbf{0.7432} & \textbf{26.87} & \textbf{0.8091} & \underline{31.29} & \textbf{0.9185} \\[0.4ex]
        \bottomrule
    \end{tabular}%
    }
\end{table*}

%% file: img_code/Figure_benchmark.tex
\begin{figure*}[!htbp]
	\scriptsize
	\centering
	\scalebox{0.89}{
    \hspace{-1.5mm}
		\begin{tabular}{lc}
            \hspace{-5mm}
            \begin{adjustbox}{valign=t}
				\begin{tabular}{c}				    
                    \includegraphics[width=0.3\textwidth, height=0.145\textheight]{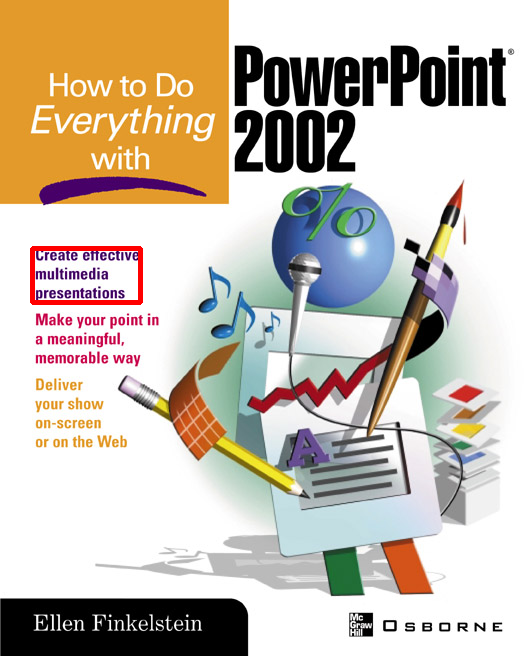} \\
					\fontsize{7.5pt}{1pt}\selectfont Set14 ($\times 2$): ppt3 \\
				\end{tabular}
			\end{adjustbox}
			\hspace{-3mm}
			\begin{adjustbox}{valign=t}
				\begin{tabular}{ccccc}
                    \includegraphics[width=0.15\textwidth, height=0.065\textheight, frame,  cfbox=black 0.5pt -0.5pt]{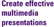} & 
					\hspace{-3mm}
					\includegraphics[width=0.15\textwidth, height=0.065\textheight, frame,  cfbox=black 0.5pt -0.5pt]{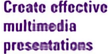} & 
					\hspace{-3mm}
					\includegraphics[width=0.15\textwidth, height=0.065\textheight, frame,  cfbox=black 0.5pt -0.5pt]{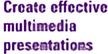} & 
                    \hspace{-3mm}
                    \includegraphics[width=0.15\textwidth, height=0.065\textheight, frame,  cfbox=black 0.5pt -0.5pt]{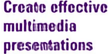} & 
                    \hspace{-3mm}
                    \includegraphics[width=0.15\textwidth, height=0.065\textheight, frame,  cfbox=black 0.5pt -0.5pt]{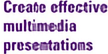}\\
					\fontsize{7.5pt}{1pt}\selectfont LR & \hspace{-2mm}
				  \fontsize{7.5pt}{1pt}\selectfont 
                    SwinIR-light~\cite{liang2021swinir} & \hspace{-2mm}
					\fontsize{7.5pt}{1pt}\selectfont SRFormer-light~\cite{zhou2023srformer} & \hspace{-2mm}
                    \fontsize{7.5pt}{1pt}\selectfont Omni-SR~\cite{wang2023omni} & \hspace{-2mm}
                    \fontsize{7.5pt}{1pt}\selectfont HiT-SIR~\cite{zhang2024hit}\\
                    \includegraphics[width=0.15\textwidth, height=0.065\textheight, frame,  cfbox=black 0.5pt -0.5pt]{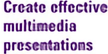} & 
                    \hspace{-3mm}
					\includegraphics[width=0.15\textwidth, height=0.065\textheight, frame,  cfbox=black 0.5pt -0.5pt]{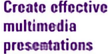} & 
					\hspace{-3mm}
					\includegraphics[width=0.15\textwidth, height=0.065\textheight, frame,  cfbox=black 0.5pt -0.5pt]{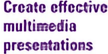} & 
					\hspace{-3mm}
                    \includegraphics[width=0.15\textwidth, height=0.065\textheight, frame,  cfbox=black 0.5pt -0.5pt]{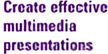} & 
					\hspace{-3mm}
					\includegraphics[width=0.15\textwidth, height=0.065\textheight, frame,  cfbox=black 0.5pt -0.5pt]{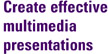} \\
                    \fontsize{7.5pt}{1pt}\selectfont ESC-lt~\cite{lee2025emulating} & \hspace{-2mm}
					\fontsize{7.5pt}{1pt}\selectfont MambaIR-light~\cite{guo2024mambair} & \hspace{-2mm}
				  \fontsize{7.5pt}{1pt}\selectfont 
                    MambaIRv2-light~\cite{guo2025mambairv2} & 
                    \hspace{-2mm}
                    \fontsize{7.5pt}{1pt}\selectfont \textbf{Ours} & 
                    \hspace{-2mm}
					\fontsize{7.5pt}{1pt}\selectfont Ground Truth \\
				\end{tabular}
			\end{adjustbox}
             \\
            \hspace{-5mm}             
            \begin{adjustbox}{valign=t}
				\begin{tabular}{c}
                \includegraphics[width=0.3\textwidth, height=0.145\textheight]{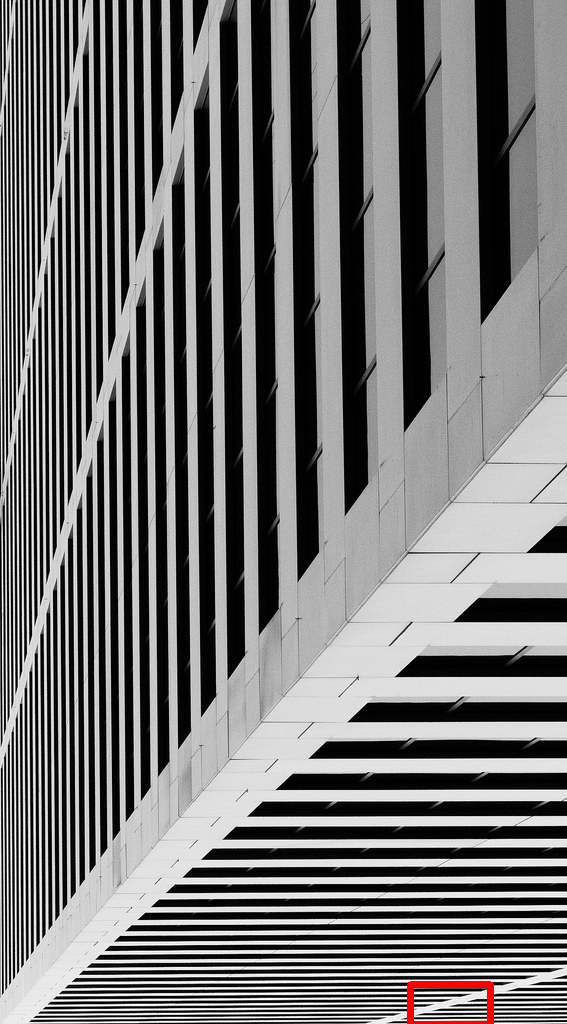} \\
					\fontsize{7.5pt}{1pt}\selectfont Urban100 ($\times 2$): img011 \\
				\end{tabular}
			\end{adjustbox}
			\hspace{-3mm}
			\begin{adjustbox}{valign=t}
				\begin{tabular}{ccccc}
                    \includegraphics[width=0.15\textwidth, height=0.065\textheight, frame,  cfbox=black 0.5pt -0.5pt]{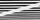} & 
					\hspace{-3mm}
					\includegraphics[width=0.15\textwidth, height=0.065\textheight, frame,  cfbox=black 0.5pt -0.5pt]{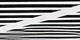} & 
					\hspace{-3mm}
					\includegraphics[width=0.15\textwidth, height=0.065\textheight, frame,  cfbox=black 0.5pt -0.5pt]{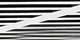} & 
                    \hspace{-3mm}
                    \includegraphics[width=0.15\textwidth, height=0.065\textheight, frame,  cfbox=black 0.5pt -0.5pt]{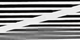} & 
                    \hspace{-3mm}
                    \includegraphics[width=0.15\textwidth, height=0.065\textheight, frame,  cfbox=black 0.5pt -0.5pt]{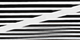}\\
					\fontsize{7.5pt}{1pt}\selectfont LR & \hspace{-2mm}
				  \fontsize{7.5pt}{1pt}\selectfont 
                    SwinIR-light~\cite{liang2021swinir} & \hspace{-2mm}
					\fontsize{7.5pt}{1pt}\selectfont SRFormer-light~\cite{zhou2023srformer} & \hspace{-2mm}
                    \fontsize{7.5pt}{1pt}\selectfont Omni-SR~\cite{wang2023omni} & \hspace{-2mm}
                    \fontsize{7.5pt}{1pt}\selectfont HiT-SIR~\cite{zhang2024hit}\\
                    \includegraphics[width=0.15\textwidth, height=0.065\textheight, frame,  cfbox=black 0.5pt -0.5pt]{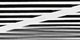} & 
                    \hspace{-3mm}
					\includegraphics[width=0.15\textwidth, height=0.065\textheight, frame,  cfbox=black 0.5pt -0.5pt]{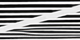} & 
					\hspace{-3mm}
					\includegraphics[width=0.15\textwidth, height=0.065\textheight, frame,  cfbox=black 0.5pt -0.5pt]{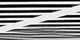} & 
					\hspace{-3mm}
                    \includegraphics[width=0.15\textwidth, height=0.065\textheight, frame,  cfbox=black 0.5pt -0.5pt]{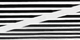} & 
					\hspace{-3mm}
					\includegraphics[width=0.15\textwidth, height=0.065\textheight, frame,  cfbox=black 0.5pt -0.5pt]{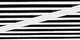} \\
                    \fontsize{7.5pt}{1pt}\selectfont ESC-lt~\cite{lee2025emulating} & \hspace{-2mm}
					\fontsize{7.5pt}{1pt}\selectfont MambaIR-light~\cite{guo2024mambair} & \hspace{-2mm}
				  \fontsize{7.5pt}{1pt}\selectfont 
                    MambaIRv2-light~\cite{guo2025mambairv2} & 
                    \hspace{-2mm}
                    \fontsize{7.5pt}{1pt}\selectfont \textbf{Ours} & 
                    \hspace{-2mm}
					\fontsize{7.5pt}{1pt}\selectfont Ground Truth \\
				\end{tabular}
			\end{adjustbox}
             \\
            \hspace{-5mm}
            \begin{adjustbox}{valign=t}
			  	\begin{tabular}{c}
			  		\includegraphics[width=0.3\textwidth, height=0.145\textheight]{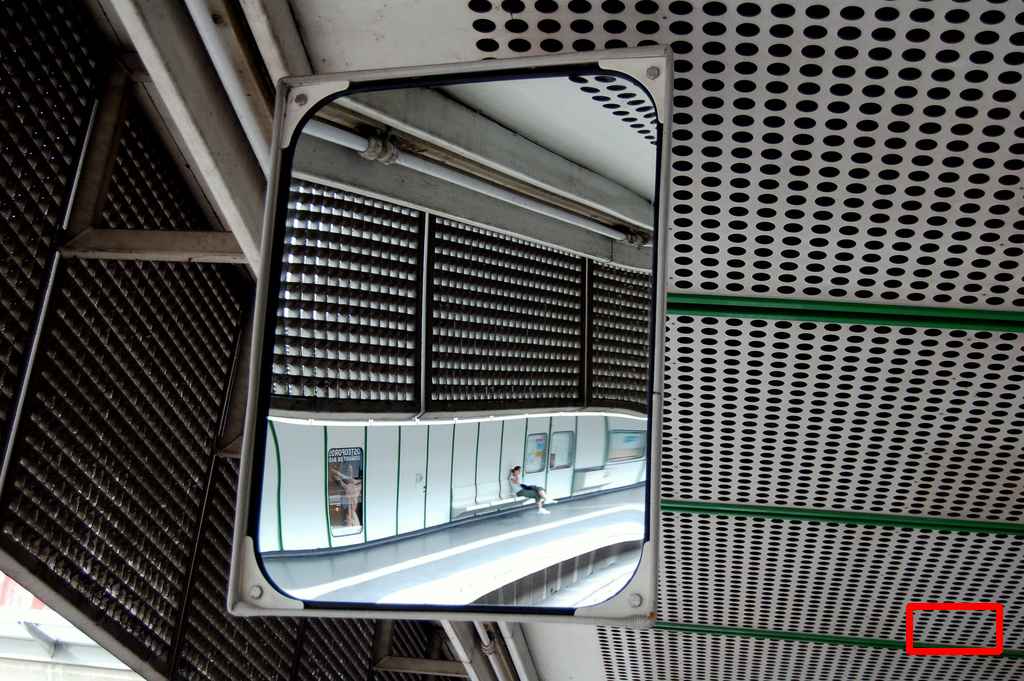} \\
			  		\fontsize{7.5pt}{1pt}\selectfont Urban100 ($\times 3$): img004 \\
			  	\end{tabular}
			  \end{adjustbox}
			  \hspace{-3mm}
			  \begin{adjustbox}{valign=t}
			 	\begin{tabular}{ccccc}
			 		\includegraphics[width=0.15\textwidth, height=0.065\textheight, frame,  cfbox=black 0.5pt -0.5pt]{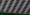} & 
			 		\hspace{-3mm}
			 		\includegraphics[width=0.15\textwidth, height=0.065\textheight, frame,  cfbox=black 0.5pt -0.5pt]{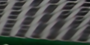} & 
			 		\hspace{-3mm}
			 		\includegraphics[width=0.15\textwidth, height=0.065\textheight, frame,  cfbox=black 0.5pt -0.5pt]{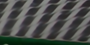} & 
                     \hspace{-3mm}
                     \includegraphics[width=0.15\textwidth, height=0.065\textheight, frame,  cfbox=black 0.5pt -0.5pt]{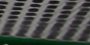} & 
                     \hspace{-3mm}
                     \includegraphics[width=0.15\textwidth, height=0.065\textheight]{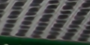}\\
			 		\fontsize{7.5pt}{1pt}\selectfont LR & \hspace{-2mm}
			 	  \fontsize{7.5pt}{1pt}\selectfont SwinIR-light~\cite{liang2021swinir} & 
                    \hspace{-2mm}
			 		\fontsize{7.5pt}{1pt}\selectfont SRFormer-light~\cite{zhou2023srformer} & \hspace{-2mm}
                    \fontsize{7.5pt}{1pt}\selectfont Omni-SR~\cite{wang2023omni} & \hspace{-2mm}
                    \fontsize{7.5pt}{1pt}\selectfont HiT-SIR~\cite{zhang2024hit}\\
                     \includegraphics[width=0.15\textwidth, height=0.065\textheight, frame,  cfbox=black 0.5pt -0.5pt]{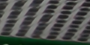} & 
                     \hspace{-3mm}
			 		\includegraphics[width=0.15\textwidth, height=0.065\textheight, frame,  cfbox=black 0.5pt -0.5pt]{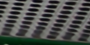} & 
					\hspace{-3mm}
			 		\includegraphics[width=0.15\textwidth, height=0.065\textheight, frame,  cfbox=black 0.5pt -0.5pt]{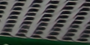} & 
			 		\hspace{-3mm}
                     \includegraphics[width=0.15\textwidth, height=0.065\textheight, frame,  cfbox=black 0.5pt -0.5pt]{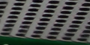} & 
			 		\hspace{-3mm}
			 		\includegraphics[width=0.15\textwidth, height=0.065\textheight, frame,  cfbox=black 0.5pt -0.5pt]{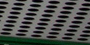} \\
                     \fontsize{7.5pt}{1pt}\selectfont ESC-lt~\cite{lee2025emulating} & \hspace{-2mm}
			 		\fontsize{7.5pt}{1pt}\selectfont MambaIR-light~\cite{guo2024mambair} & \hspace{-2mm}
			 	  \fontsize{7.5pt}{1pt}\selectfont 
                     MambaIRv2-light~\cite{guo2025mambairv2} & 
                     \hspace{-2mm}
                     \fontsize{7.5pt}{1pt}\selectfont \textbf{Ours} & 
                     \hspace{-2mm}
			 		\fontsize{7.5pt}{1pt}\selectfont Ground Truth \\
			 	\end{tabular}
			 \end{adjustbox}
              \\
            \hspace{-5mm}
             \begin{adjustbox}{valign=t}
				\begin{tabular}{c}
					\includegraphics[width=0.3\textwidth, height=0.145\textheight]{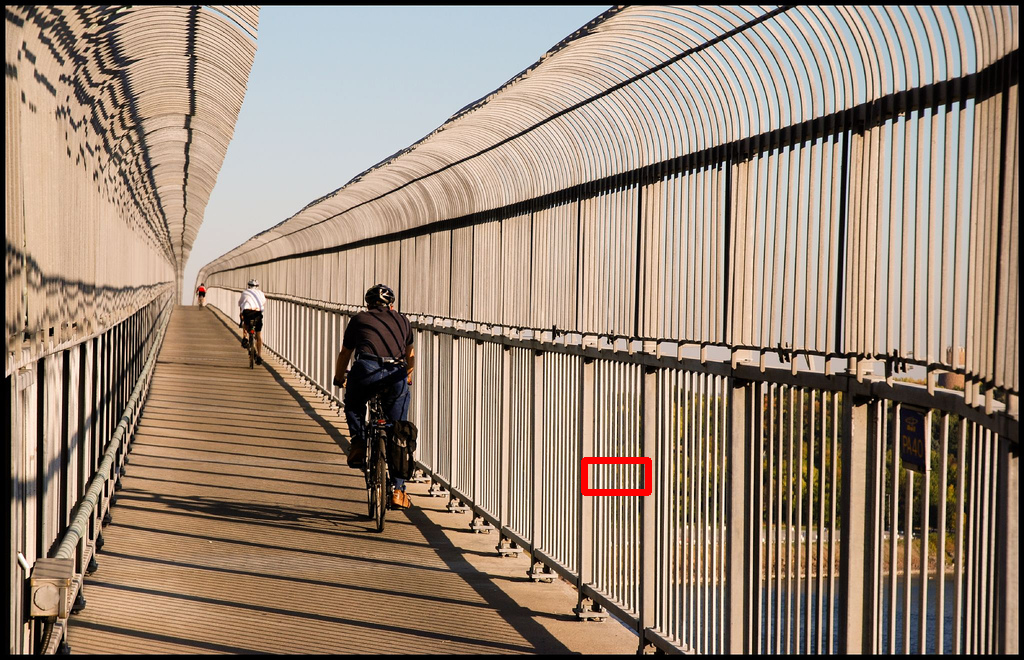} \\
					\fontsize{7.5pt}{1pt}\selectfont Urban100 ($\times 4$): img024 \\
				\end{tabular}
			\end{adjustbox}
			\hspace{-3mm}
			\begin{adjustbox}{valign=t}
				\begin{tabular}{ccccc}
					\includegraphics[width=0.15\textwidth, height=0.065\textheight, frame,  cfbox=black 0.5pt -0.5pt]{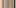} & 
					\hspace{-3mm}
					\includegraphics[width=0.15\textwidth, height=0.065\textheight, frame,  cfbox=black 0.5pt -0.5pt]{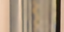} & 
					\hspace{-3mm}
					\includegraphics[width=0.15\textwidth, height=0.065\textheight, frame,  cfbox=black 0.5pt -0.5pt]{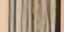} & 
                    \hspace{-3mm}
                    \includegraphics[width=0.15\textwidth, height=0.065\textheight, frame,  cfbox=black 0.5pt -0.5pt]{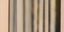} & 
                    \hspace{-3mm}
                    \includegraphics[width=0.15\textwidth, height=0.065\textheight, frame,  cfbox=black 0.5pt -0.5pt]{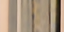}\\
					\fontsize{7.5pt}{1pt}\selectfont LR & \hspace{-2mm}
				  \fontsize{7.5pt}{1pt}\selectfont SwinIR-light~\cite{liang2021swinir} & 
                    \hspace{-2mm}
					\fontsize{7.5pt}{1pt}\selectfont SRFormer-light~\cite{zhou2023srformer} & \hspace{-2mm}
                    \fontsize{7.5pt}{1pt}\selectfont Omni-SR~\cite{wang2023omni} & \hspace{-2mm}
                    \fontsize{7.5pt}{1pt}\selectfont HiT-SIR~\cite{zhang2024hit}\\
                    \includegraphics[width=0.15\textwidth, height=0.065\textheight, frame,  cfbox=black 0.5pt -0.5pt]{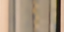} & 
                    \hspace{-3mm}
					\includegraphics[width=0.15\textwidth, height=0.065\textheight, frame,  cfbox=black 0.5pt -0.5pt]{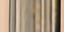} & 
					\hspace{-3mm}
					\includegraphics[width=0.15\textwidth, height=0.065\textheight, frame,  cfbox=black 0.5pt -0.5pt]{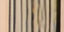} & 
					\hspace{-3mm}
                    \includegraphics[width=0.15\textwidth, height=0.065\textheight, frame,  cfbox=black 0.5pt -0.5pt]{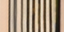} & 
					\hspace{-3mm}
					\includegraphics[width=0.15\textwidth, height=0.065\textheight, frame,  cfbox=black 0.5pt -0.5pt]{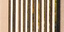} \\
                    \fontsize{7.5pt}{1pt}\selectfont ESC-lt~\cite{lee2025emulating} & \hspace{-2mm}
					\fontsize{7.5pt}{1pt}\selectfont MambaIR-light~\cite{guo2024mambair} & \hspace{-2mm}
				  \fontsize{7.5pt}{1pt}\selectfont MambaIRv2-light~\cite{guo2025mambairv2} & 
                    \hspace{-2mm}
                    \fontsize{7.5pt}{1pt}\selectfont \textbf{Ours} & 
                    \hspace{-2mm}
					\fontsize{7.5pt}{1pt}\selectfont Ground Truth \\
				\end{tabular}
			\end{adjustbox}
             \\
	  \end{tabular}} 
	\vspace{-2mm}
	\caption{Qualitative comparisons with existing methods in different scenes. Our method can restore clearer edges and structures.}
    \vspace{-3mm}
	\label{visual_benchmark}
\end{figure*}

%% file: img_code/Figure_RealSR.tex
\begin{figure*}[!t]
    \scriptsize
    \centering
    \scalebox{0.81}{
    \hspace{-5.5mm}
    \begin{tabular}{lc}
        \begin{adjustbox}{valign=t}
	\begin{tabular}{c}	 
            \includegraphics[width=0.25\textwidth,height=0.125\textheight]{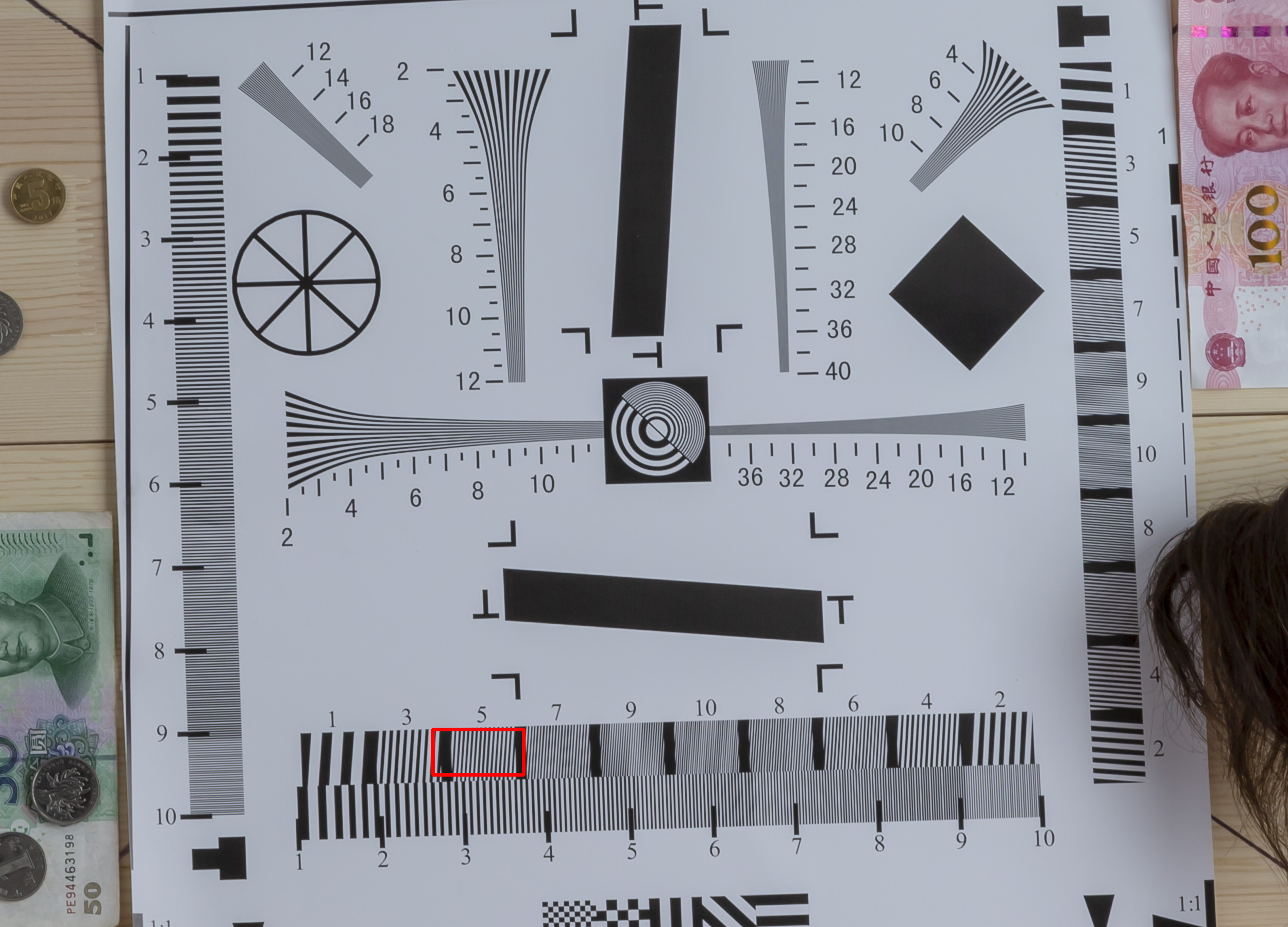} \\
		\fontsize{6.5pt}{1pt}\selectfont RealSRv3 ($\times 4$): Canon045 \\
	\end{tabular}
	\end{adjustbox}
	\hspace{-4mm}
	\begin{adjustbox}{valign=t}
	\begin{tabular}{ccc}
		\includegraphics[width=0.11\textwidth, height=0.055\textheight, frame,  cfbox=black 0.5pt -0.5pt]{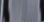} & 
		\hspace{-4mm}
            \includegraphics[width=0.11\textwidth, height=0.055\textheight, frame,  cfbox=black 0.5pt -0.5pt]{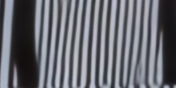} & 
		\hspace{-4mm}
		\includegraphics[width=0.11\textwidth, height=0.055\textheight, frame,  cfbox=black 0.5pt -0.5pt]{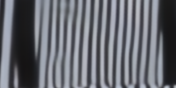} \\
		\fontsize{6.5pt}{1pt}\selectfont LR & \hspace{-3mm}
		\fontsize{6.5pt}{1pt}\selectfont HiT-SIR~\cite{zhang2024hit} & \hspace{-4mm}
            \fontsize{6.5pt}{1pt}\selectfont MambaIR-light~\cite{guo2024mambair}\\			
            \includegraphics[width=0.11\textwidth, height=0.055\textheight, frame,  cfbox=black 0.5pt -0.5pt]{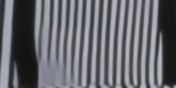} & 
		\hspace{-4mm}
            \includegraphics[width=0.11\textwidth, height=0.055\textheight, frame,  cfbox=black 0.5pt -0.5pt]{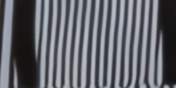} & 
		\hspace{-4mm}
		\includegraphics[width=0.11\textwidth, height=0.055\textheight, frame,  cfbox=black 0.5pt -0.5pt]{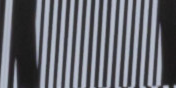} \\
		\fontsize{6.5pt}{1pt}\selectfont MambaIRv2-light~\cite{guo2025mambairv2} & \hspace{-4mm}
            \fontsize{6.5pt}{1pt}\selectfont \textbf{Ours} & \hspace{-4mm}
		\fontsize{6.5pt}{1pt}\selectfont Ground Truth \\
        \end{tabular}
	\end{adjustbox}
        \hspace{-4mm}
        \begin{adjustbox}{valign=t}
	\begin{tabular}{c}
	    \includegraphics[width=0.25\textwidth, height=0.125\textheight]{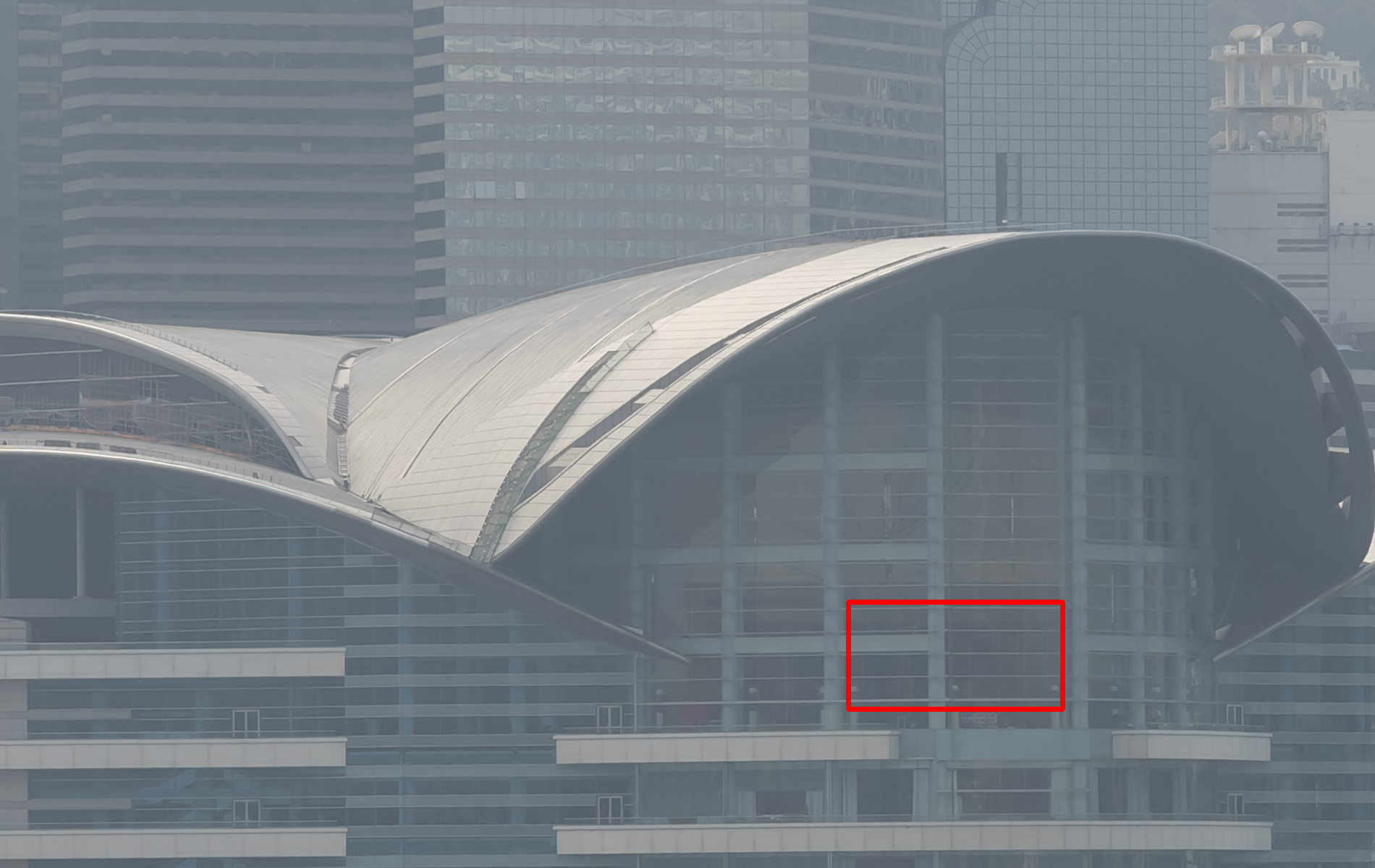} \\
	    \fontsize{6.5pt}{1pt}\selectfont RealSRv3 ($\times 4$): Nikon042 \\
	\end{tabular}
	\end{adjustbox}
	\hspace{-4mm}
	\begin{adjustbox}{valign=t}
	\begin{tabular}{ccc}
		\includegraphics[width=0.11\textwidth, height=0.055\textheight, frame,  cfbox=black 0.5pt -0.5pt]{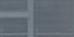} & 
		\hspace{-4mm}
            \includegraphics[width=0.11\textwidth, height=0.055\textheight, frame,  cfbox=black 0.5pt -0.5pt]{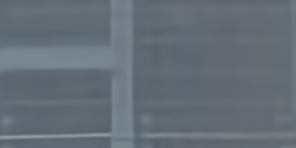} & 
		\hspace{-4mm}
		\includegraphics[width=0.11\textwidth, height=0.055\textheight, frame,  cfbox=black 0.5pt -0.5pt]{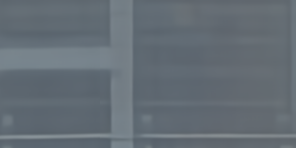} \\
		\fontsize{6.5pt}{1pt}\selectfont LR & \hspace{-3mm}
		\fontsize{6.5pt}{1pt}\selectfont HiT-SIR~\cite{zhang2024hit} & \hspace{-4mm}
            \fontsize{6.5pt}{1pt}\selectfont MambaIR-light~\cite{guo2024mambair}\\	
		\includegraphics[width=0.11\textwidth, height=0.055\textheight, frame,  cfbox=black 0.5pt -0.5pt]{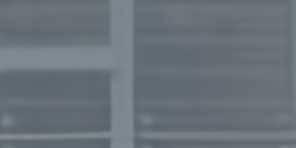} & 
		\hspace{-4mm}
            \includegraphics[width=0.11\textwidth, height=0.055\textheight, frame,  cfbox=black 0.5pt -0.5pt]{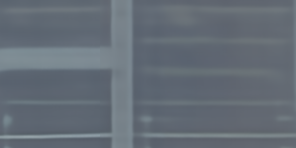} & 
		\hspace{-4mm}
		\includegraphics[width=0.11\textwidth, height=0.055\textheight, frame,  cfbox=black 0.5pt -0.5pt]{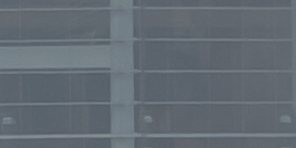} \\
            \fontsize{6.5pt}{1pt}\selectfont MambaIRv2-light~\cite{guo2025mambairv2} & \hspace{-3mm}
            \fontsize{6.5pt}{1pt}\selectfont \textbf{Ours} & \hspace{-4mm}
		\fontsize{6.5pt}{1pt}\selectfont Ground Truth \\
	\end{tabular}
	  \end{adjustbox}
	\\
   \end{tabular} }
   \vspace{-2mm}            
   \caption{Qualitative comparisons of our T-PMambaSR with existing methods on the real-world test set RealSRv3~\cite{cai2019toward}.}
   \label{visual_realsr}
\end{figure*}

%% file: img_code/Figure_receptive_field.tex
\begin{figure*}[!t]
	\scriptsize
	\centering
	\scalebox{0.96}{
    \hspace{-5.2mm}
		\begin{tabular}{c}
            \begin{adjustbox}{valign=t}
				\begin{tabular}{c}				    
                    \includegraphics[width=67mm, height=78mm]{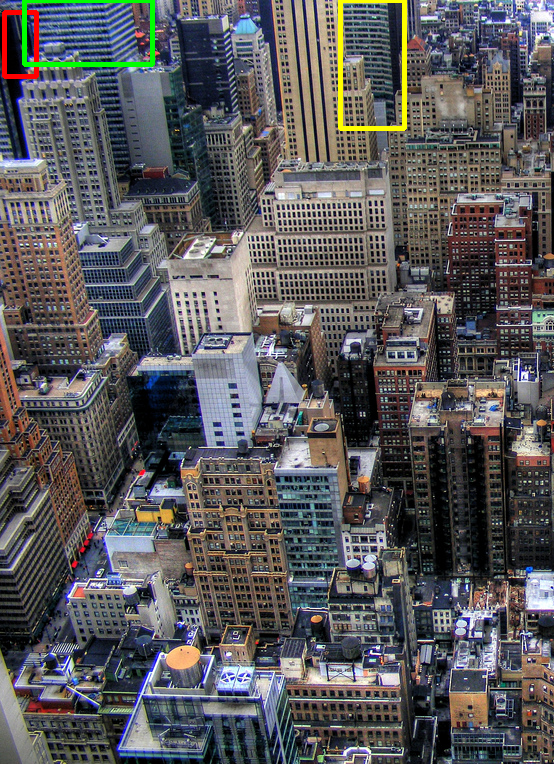} 
                    \\
					\fontsize{7.5pt}{1pt}\selectfont Urban100 ($\times 4$): img073 
                    \\
				\end{tabular}
			\end{adjustbox}
			\hspace{-4mm}
			\begin{adjustbox}{valign=t}
				\begin{tabular}{llllllll}
                    \includegraphics[width=0.0675\textwidth, height=0.08\textheight, frame,  cfbox=black 0.5pt -0.5pt]{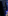}  
					\hspace{1mm}
                    
					\includegraphics[width=0.0675\textwidth, height=0.08\textheight, frame,  cfbox=black 0.5pt -0.5pt]{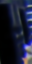} 
					\hspace{1mm}
                    
					\includegraphics[width=0.0675\textwidth, height=0.08\textheight, frame,  cfbox=black 0.5pt -0.5pt]{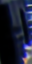} 
                    \hspace{1mm}
                    
                    \includegraphics[width=0.0675\textwidth, height=0.08\textheight, frame,  cfbox=black 0.5pt -0.5pt]{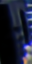} 
                    \hspace{1mm}
                    
                    \includegraphics[width=0.0675\textwidth, height=0.08\textheight, frame,  cfbox=black 0.5pt -0.5pt]{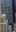}
                    \hspace{1mm}

                    \includegraphics[width=0.0675\textwidth, height=0.08\textheight, frame,  cfbox=black 0.5pt -0.5pt]{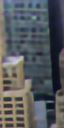}
                    \hspace{1mm}

                    \includegraphics[width=0.0675\textwidth, height=0.08\textheight, frame,  cfbox=black 0.5pt -0.5pt]{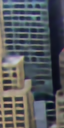}
                    \hspace{1mm}

                    \includegraphics[width=0.0675\textwidth, height=0.08\textheight, frame,  cfbox=black 0.5pt -0.5pt]{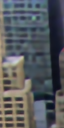}
                    
                    \\
                    \hspace{-0.5mm}
					\fontsize{7.5pt}{1pt}\selectfont Patch1-LR
                    \hspace{0.5mm}

					\fontsize{7.5pt}{1pt}\selectfont SRFormer~\cite{zhou2023srformer}  \hspace{-1mm}
                    
                    \fontsize{7.5pt}{1pt}\selectfont HiT-SIR~\cite{zhang2024hit} \hspace{1mm}
                    
                    \fontsize{7.5pt}{1pt}\selectfont ESC-lt~\cite{lee2025emulating} 
                    \hspace{1.5mm}
                    
                    \fontsize{7.5pt}{1pt}\selectfont Patch2-LR 
                    \hspace{0.5mm}

					\fontsize{7.5pt}{1pt}\selectfont SRFormer~\cite{zhou2023srformer} 
                    \hspace{-1mm}
                    
                    \fontsize{7.5pt}{1pt}\selectfont HiT-SIR~\cite{zhang2024hit}  \hspace{1mm}
                    
                    \fontsize{7.5pt}{1pt}\selectfont ESC-lt~\cite{lee2025emulating} 
                    
                    \\
                    
                    \includegraphics[width=0.0675\textwidth, height=0.08\textheight, frame,  cfbox=black 0.5pt -0.5pt]{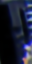}  
					\hspace{1mm}
                    
					\includegraphics[width=0.0675\textwidth, height=0.08\textheight, frame,  cfbox=black 0.5pt -0.5pt]{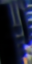} 
					\hspace{1mm}
                    
					\includegraphics[width=0.0675\textwidth, height=0.08\textheight, frame,  cfbox=black 0.5pt -0.5pt]{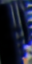} 
                    \hspace{1mm}
                    
                    \includegraphics[width=0.0675\textwidth, height=0.08\textheight, frame,  cfbox=black 0.5pt -0.5pt]{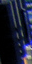}  
                    \hspace{1mm}
                                        
                    \includegraphics[width=0.0675\textwidth, height=0.08\textheight, frame,  cfbox=black 0.5pt -0.5pt]{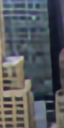}
                    \hspace{1mm}

                    \includegraphics[width=0.0675\textwidth, height=0.08\textheight, frame,  cfbox=black 0.5pt -0.5pt]{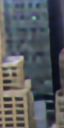}
                    \hspace{1mm}

                    \includegraphics[width=0.0675\textwidth, height=0.08\textheight, frame,  cfbox=black 0.5pt -0.5pt]{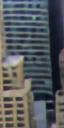}
                    \hspace{1mm}

                    \includegraphics[width=0.0675\textwidth, height=0.08\textheight, frame,  cfbox=black 0.5pt -0.5pt]{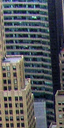}
                    
                    \\
                    \hspace{-2mm}
					\fontsize{7.5pt}{1pt}\selectfont MambaIR~\cite{guo2024mambair}  
                    \hspace{-1mm}

					\fontsize{7.5pt}{1pt}\selectfont MambaIRv2~\cite{guo2025mambairv2}  
                    \hspace{0mm}
                    
                    \fontsize{7.5pt}{1pt}\selectfont \textbf{Ours} 
                    \hspace{8.5mm}
                    
                    \fontsize{7.5pt}{1pt}\selectfont GT 
                    \hspace{4mm}
                    
                    \fontsize{7.5pt}{1pt}\selectfont MambaIR~\cite{guo2024mambair} 
                    \hspace{-1mm}

					\fontsize{7.5pt}{1pt}\selectfont MambaIRv2~\cite{guo2025mambairv2} 
                    \hspace{0.5mm}
                    
                    \fontsize{7.5pt}{1pt}\selectfont \textbf{Ours}  
                    \hspace{9.5mm}
                    
                    \fontsize{7.5pt}{1pt}\selectfont GT  
                    
                    \\
                                        
                    \includegraphics[width=0.15\textwidth, frame,  cfbox=black 0.5pt -0.5pt]{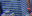}
                    \hspace{1.2mm}

                    \includegraphics[width=0.15\textwidth, frame,  cfbox=black 0.5pt -0.5pt]{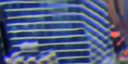}
                    \hspace{1.3mm}
                    
                    \includegraphics[width=0.15\textwidth, frame,  cfbox=black 0.5pt -0.5pt]{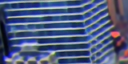}
                    \hspace{1.4mm}

                    \includegraphics[width=0.15\textwidth, frame,  cfbox=black 0.5pt -0.5pt]{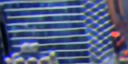}
                    
                    \\
                    \hspace{7.5mm}
                    \fontsize{7.5pt}{1pt}\selectfont Patch3-LR 
                    \hspace{15mm}

					\fontsize{7.5pt}{1pt}\selectfont SRFormer~\cite{zhou2023srformer} 
                    \hspace{14mm}
                    
                    \fontsize{7.5pt}{1pt}\selectfont HiT-SIR~\cite{zhang2024hit} \hspace{17mm}
                    
                    \fontsize{7.5pt}{1pt}\selectfont ESC-lt~\cite{lee2025emulating} 

                    \\
                    
                    \includegraphics[width=0.15\textwidth, frame,  cfbox=black 0.5pt -0.5pt]{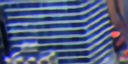} 
                    \hspace{1.2mm}

                    \includegraphics[width=0.15\textwidth, frame,  cfbox=black 0.5pt -0.5pt]{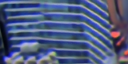}
                    \hspace{1.3mm}
                    
                    \includegraphics[width=0.15\textwidth, frame,  cfbox=black 0.5pt -0.5pt]{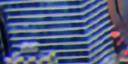} 
                    \hspace{1.4mm}

                    \includegraphics[width=0.15\textwidth, frame,  cfbox=black 0.5pt -0.5pt]{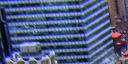}
                    \\
                    \hspace{5mm}
                    \fontsize{7.5pt}{1pt}\selectfont MambaIR~\cite{guo2024mambair} 
                    \hspace{12mm}

					\fontsize{7.5pt}{1pt}\selectfont MambaIRv2~\cite{guo2025mambairv2} 
                    \hspace{17mm}
                    
                    \fontsize{7.5pt}{1pt}\selectfont \textbf{Ours} 
                    \hspace{25mm}
                    
                    \fontsize{7.5pt}{1pt}\selectfont GT
				\end{tabular}
			\end{adjustbox}
             \\
	  \end{tabular}}

\vspace{0mm}
\caption{Visual comparisons of different SR methods on multi-scale targets. The target in \textbf{Patch 1} represents local information, while the targets in \textbf{Patch 2} and \textbf{Patch 3} exceed the standard Window-MHSA with a size of $16\times16$. Competing Transformer and Mamba architectures exhibit mixed performance across these varying scales. In contrast, our method's hybrid architecture and progressive receptive field expansion strategy consistently yields the best visual quality across all targets.}
\label{visual_receptive_field}
\vspace{-2mm}
\end{figure*}

%% file: img_code/Figure_LAM.tex
\begin{figure*}[!htbp]
\vspace{3.5mm}
\begin{overpic}[width=0.999\linewidth]{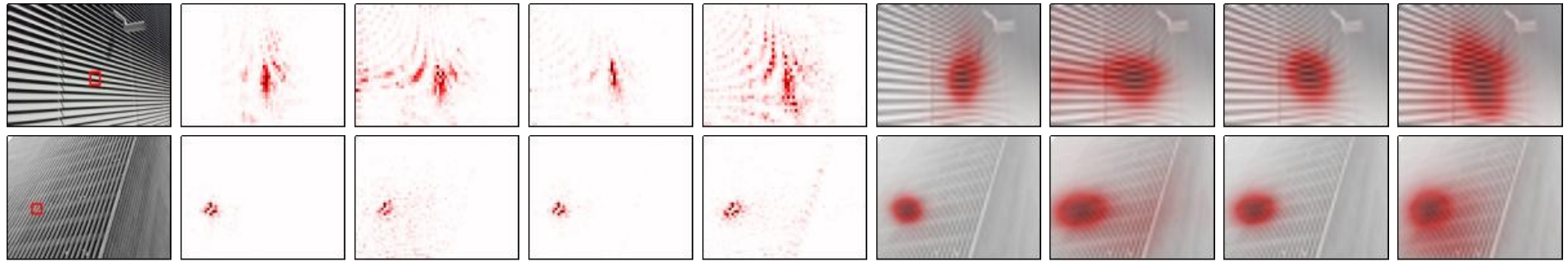}
\put(27.5,17.15){\color{black}{\fontsize{8pt}{1pt}\selectfont LAM Attribution}}
\put(70.5,17.15){\color{black}{\fontsize{8pt}{1pt}\selectfont Pixel Contribution Area}}

\put(2.2,-1.2){\color{black}{\fontsize{6.5pt}{1pt}\selectfont Ground Truth}}
\put(11.63,-1.2){\color{black}{\fontsize{6.5pt}{1pt}\selectfont SRFormer-light~\cite{zhou2023srformer}}}
\put(22.75,-1.2){\color{black}{\fontsize{6.5pt}{1pt}\selectfont MambaIR-light~\cite{guo2024mambair}}}
\put(33.35,-1.2){\color{black}{\fontsize{6.5pt}{1pt}\selectfont MambaIRv2-light~\cite{guo2025mambairv2}}}
\put(48.6,-1.2){\color{black}{\fontsize{6.5pt}{1pt}\selectfont \bf{Ours}}}

\put(56,-1.2){\color{black}{\fontsize{6.5pt}{1pt}\selectfont SRFormer-light~\cite{zhou2023srformer}}}
\put(67.15,-1.2){\color{black}{\fontsize{6.5pt}{1pt}\selectfont MambaIR-light~\cite{guo2024mambair}}}
\put(77.75,-1.2){\color{black}{\fontsize{6.5pt}{1pt}\selectfont MambaIRv2-light~\cite{guo2025mambairv2}}}
\put(93.1,-1.2){\color{black}{\fontsize{6.5pt}{1pt}\selectfont \bf{Ours}}}

\put(12.0,0.6){\color{black}{\fontsize{6.5pt}{1pt}\selectfont DI: 5.939}}
\put(23.2,0.6){\color{black}{\fontsize{6.5pt}{1pt}\selectfont DI: 19.286}}
\put(34.35,0.6){\color{black}{\fontsize{6.5pt}{1pt}\selectfont DI: 9.213}}
\put(45.5,0.6){\color{black}{\fontsize{6.5pt}{1pt}\selectfont \bf{DI: 20.017}}}

\put(12.0,9.0){\color{black}{\fontsize{6.5pt}{1pt}\selectfont DI: 16.319}}
\put(23.2,9.0){\color{black}{\fontsize{6.5pt}{1pt}\selectfont DI: 22.183}}
\put(34.35,9.0){\color{black}{\fontsize{6.5pt}{1pt}\selectfont DI: 18.391}}
\put(45.5,9.0){\color{black}{\fontsize{6.5pt}{1pt}\selectfont \bf{DI: 24.641}}}

\end{overpic}
\vspace{-2mm}
   \caption{LAM~\cite{gu2021interpreting} comparison between our model and existing methods~\cite{zhou2023srformer,guo2024mambair,guo2025mambairv2} on the $\times4$ scale. Our method leads to the maximum pixel contribution areas, confirming its superior capacity for capturing broad contextual information.}
\label{fig:LAM}
\vspace{-2mm}
\end{figure*}